\documentclass[10pt,twocolumn,letterpaper]{article}

\usepackage[pagenumbers]{cvpr} %

\usepackage{cuted}   %
\usepackage{caption} %

\newsavebox{\facetablebox}

\usepackage{booktabs}
\usepackage{multirow}
\usepackage{graphicx}
\usepackage{listings}
\usepackage{makecell}
\usepackage{tabularx}
\usepackage{colortbl}
\usepackage{array}
\usepackage{amssymb}
\usepackage{pifont}
\usepackage{float}
\usepackage{caption}
\usepackage{subcaption}
\newsavebox{\ocaltablebox}
\ifdefined\facetablebox\else\newsavebox{\facetablebox}\fi %

\definecolor{linkblue}{rgb}{0.21,0.49,0.74}   %
\definecolor{figgreen}{RGB}{79,98,40}        %
\usepackage[pagebackref,breaklinks,colorlinks,allcolors=linkblue]{hyperref}

\definecolor{markgreen}{RGB}{34,139,34}
\definecolor{markred}{RGB}{200,40,40}
\newcommand{\cmark}{\textcolor{markgreen}{\ding{51}}}
\newcommand{\xmark}{\textcolor{markred}{\ding{55}}}

\usepackage{xspace}
\usepackage{longtable}

\newcommand{\nHours}{10{,}000\xspace}
\newcommand{\nEpisodes}{1{,}378{,}462\xspace}
\newcommand{\nTimesteps}{$1.08\times10^9$\xspace}
\newcommand{\nFrames}{$6.5\times10^9$\xspace}

\newcommand{\nCountries}{11\xspace}
\newcommand{\nContinents}{five\xspace}
\newcommand{\nVerticals}{18\xspace}
\newcommand{\nSiteTypes}{200\xspace}
\newcommand{\nTasks}{2{,}000\xspace}
\newcommand{\nTasksEasy}{769\xspace}
\newcommand{\nTasksMedium}{761\xspace}
\newcommand{\nTasksHard}{470\xspace}
\newcommand{\nTasksTopFour}{699\xspace}

\title{RoboCap-10k: A Metrically Accurate Egocentric Dataset for Robot Learning}

\author{%
Kheri Hughes\textsuperscript{1} \quad
Qasim Ali\textsuperscript{1} \quad
Nikash Bhardwaj\textsuperscript{1} \quad
Owen Akhibi\textsuperscript{1} \quad \\
Niresh Dravin\textsuperscript{2} \quad
Vivek Sharma\textsuperscript{2} \quad
Sam Cho\textsuperscript{2} \quad
Ethan Lee\textsuperscript{2} \quad
Paul Luo\textsuperscript{2} \quad
Sunny Yu\textsuperscript{2} \quad \\
Jonathan Victor\textsuperscript{2} \quad
Amanda Young\textsuperscript{2} \quad
Michael Cho\textsuperscript{2} \quad
Ademi Adeniji\textsuperscript{1} \quad
Vincent Liu\textsuperscript{1} \\[0.6em]
\textsuperscript{1}Grounded Superintelligence \quad
\textsuperscript{2}BitRobot \\[0.6em]
\href{https://www.gsi.company}{\texttt{gsi.company}}
}

\begin{document}

\twocolumn[{%
\renewcommand\twocolumn[1][]{#1}%
\maketitle
\begin{center}
  \centering
  \captionsetup{type=figure}%
  \includegraphics[width=\linewidth]{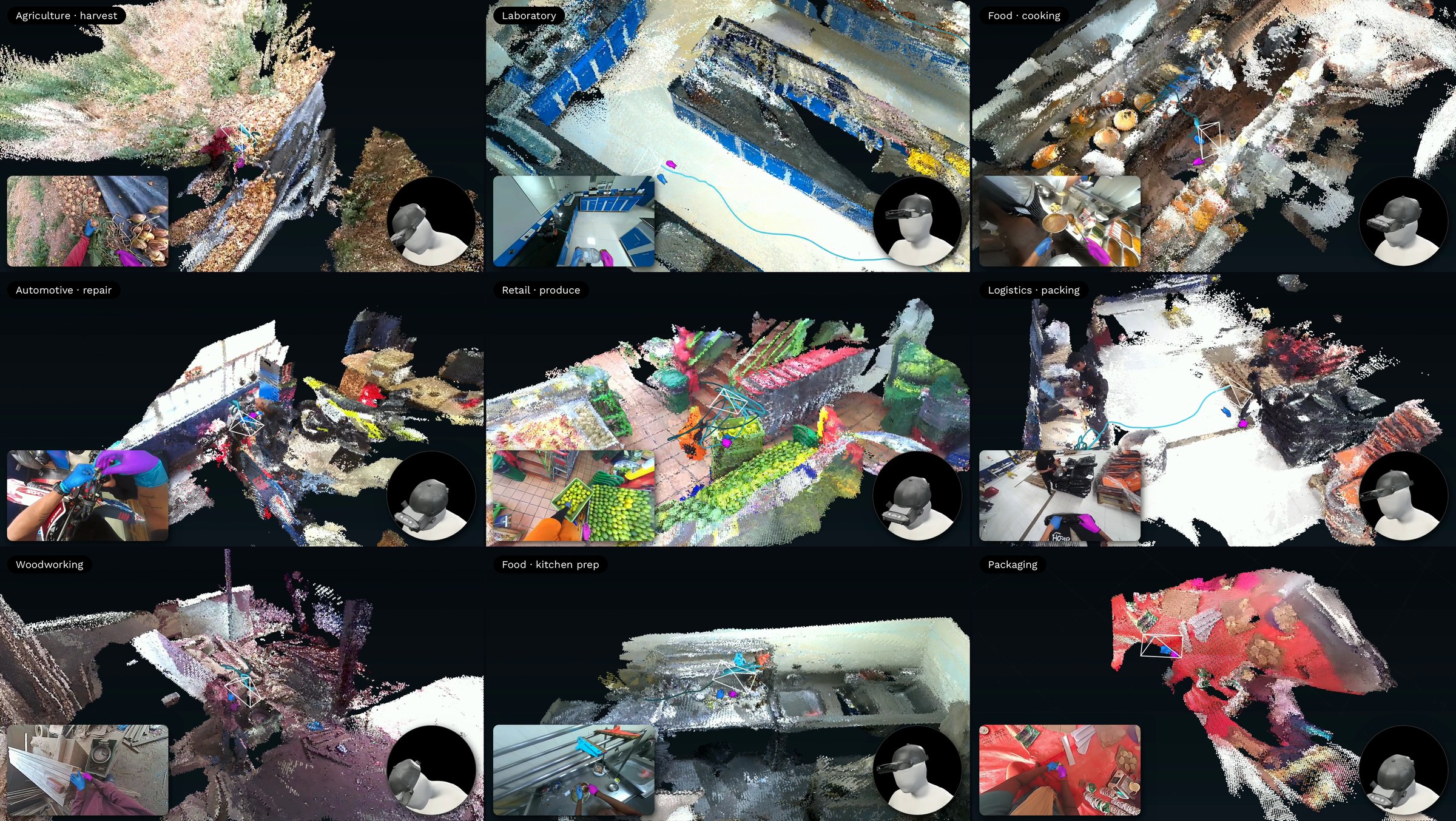}
  \caption{\textbf{RoboCap-10k}: \nHours hours of egocentric occupational work, captured on calibrated RoboCap devices with metric 3D head and hand poses, and annotated with task and sub-task language. Insets: egocentric view (bottom left) and head pose (bottom right).}
  \label{fig:teaser}
  \vspace{1.25em}
\end{center}
}]

\begin{abstract}
Egocentric video is the most scalable source of manipulation data for robot learning, but its difficulty lies in producing high-quality supervision. Existing large corpora lack metric 3D labels or rely on heterogeneous devices, motion-capture labs, or a headset's built-in real-time tracker, whose accuracy is not reported for the released data. We introduce RoboCap-10k: \nHours curated hours of unscripted manipulation recorded during workers' paid shifts in \nCountries countries, spanning \nVerticals task families, \nSiteTypes task types, and \nTasks tasks. Every hour is captured on one device, the RoboCap, a 250\,g hat with six synchronized global-shutter cameras and dual IMUs. Online calibration brings every device below 1\,px of epipolar error, visual-inertial SLAM yields gravity-aligned 6-DoF head trajectories, and a multi-view hand tracker yields metric cm-precise hand poses. Each episode is annotated with task and sub-task language captions. Project page: \href{https://robocap10k.com}{\texttt{https://robocap10k.com}}.
\end{abstract}

\section{Introduction}
\label{sec:intro}

The pursuit of general-purpose robotics has exposed a data bottleneck: whereas foundation models for language and vision scale on internet text and imagery, robot policies remain starved for physical interaction data. Teleoperation provides demonstrations in the robot's own embodiment, but it is slow, expensive, and hard to extend to the long tail of environments and tasks. The field has therefore turned to egocentric human video, which captures fluid, purposeful manipulation across thousands of real environments at a fraction of the cost.

Egocentric video becomes action supervision only through estimation. Recent scaling studies show that pretraining on large egocentric corpora improves generalization of robot policies, and that 3D hand pose labels are needed to extract most of that benefit~\cite{ma2026humanscale,egoscale2026,sun2026ego4wam}. The quality of those labels is set by the capture hardware. The largest corpora are recorded with monocular head-mounted cameras~\cite{grauman2022ego4d,buildai2025egocentric10k}, from which metric scale can only be inferred from learned priors. Calibrated stereo rigs measure scale directly, but are sensitive to calibration: on the RoboCap, one pixel of disparity error corresponds to roughly one centimeter of depth at 0.65\,m, comparable to the grasp tolerance of a parallel-jaw gripper~\cite{ali2026robocapnewplatformegocentric}. Research headsets paired with optical motion capture reach millimeter accuracy, but only within instrumented volumes and with small participant pools~\cite{banerjee2024hot3d,han2022umetrack}. Federated efforts scale by pooling heterogeneous devices~\cite{punamiya2026egoverse}, which introduces cross-rig variation in both appearance and 3D estimation quality. Datasets labeled by a headset's own tracker inherit whatever that tracker outputs in real time, with no raw multi-view data to re-estimate from. RoboCap-10k instead releases the raw synchronized streams alongside offline labels whose accuracy is benchmarked (Section~\ref{sec:experiments}), so labels can be audited and regenerated as methods improve.

RoboCap-10k also standardizes the hardware: it contains a single device model collecting real paid work processed by a single 3D stack. All \nHours hours are recorded on the RoboCap~\cite{ali2026robocapnewplatformegocentric}, a 250\,g hat whose camera layout is designed based on the requirements of near-field metric estimation, and all are processed by the same Grounded API pipeline with per-session online calibration. Rather than scripted household chores or tabletop tasks, the corpus records occupational work, including bread and pastry production, industrial garment stitching, engine and motorcycle repair, circuit-board soldering, rebar tying, and onion harvesting, in workplaces across India, Nigeria, Colombia, the United States, the United Kingdom, Mexico, Chile, Malaysia, the Philippines, Indonesia, and Spain.

Our contributions are:

\begin{enumerate}
    \item \textbf{A \nHours-hour curated, single-device corpus of occupational manipulation.} Every hour is bimanual manipulation with a clear task and passes SLAM and hand-tracking quality checks. RoboCap-10k covers \nVerticals task families, \nSiteTypes task types, and \nTasks distinct tasks in \nCountries countries across \nContinents continents. Because every hour comes from the same hardware design with per-unit factory calibration, there is no cross-device domain shift in either imagery or 3D label quality.
    \item \textbf{Metric 3D annotations for every frame.} Online-calibrated, gravity-aligned 6-DoF head trajectories from multi-camera visual-inertial SLAM and metric MANO hand poses from a multi-view hand tracker, expressed in the head-mounted rig frame and mapped to the world frame by the head trajectory.
    \item \textbf{A two-tier language layer.} An episode-level caption (scene and high-level task) paired with dense, temporally segmented sub-task units, each carrying a high-level task and a low-level action.
\end{enumerate}

\begin{table*}[t]
\centering
\scriptsize
\setlength{\tabcolsep}{3pt}
\begin{tabular*}{\textwidth}{@{\extracolsep{\fill}}lrlccccccl@{}}
\toprule
\textbf{Dataset} & \makecell{\textbf{Released}\\\textbf{ego hours}} & \textbf{Capture device} & \makecell{\textbf{Single}\\\textbf{device}} & \makecell{\textbf{Calibrated}\\\textbf{stereo}} & \makecell{\textbf{Metric}\\\textbf{6-DoF head}} & \makecell{\textbf{3D}\\\textbf{hands}} & \textbf{Language} & \makecell{\textbf{Occupational}\\\textbf{work}} & \textbf{Setting} \\
\midrule
EPIC-KITCHENS-100~\cite{damen2018epickitchens} & 100 & GoPro & \cmark & \xmark & \xmark & \xmark & \cmark & \xmark & Home kitchens \\
Ego4D~\cite{grauman2022ego4d} & 3,670 & Mixed consumer cameras & \xmark & \xmark & \xmark & \xmark & \cmark & \xmark & Daily life \\
Ego-Exo4D~\cite{grauman2024egoexo4d} & 221 & Aria + exo GoPros & \cmark & \xmark & \cmark & \xmark & \cmark & \xmark & Skilled activities \\
Nymeria~\cite{ma2024nymeria} & 300 & Aria + XSens & \cmark & \xmark & \cmark & \xmark & \cmark & \xmark & Daily life \\
HOT3D~\cite{banerjee2024hot3d} & 14 & Aria, Quest 3 & \xmark & \xmark & \cmark & \cmark & \xmark & \xmark & Lab (mocap) \\
EgoDex~\cite{hoque2025egodex} & 829 & Apple Vision Pro & \cmark & \xmark & \cmark & \cmark & \cmark & \xmark & Tabletop \\
EgoVerse~\cite{punamiya2026egoverse} & 1,362 & Aria, iPhone, stereo rigs & \xmark & \xmark & \cmark & \cmark & \cmark & \xmark & Mixed \\
EgoLive~\cite{li2026egolive} & 1,680 & Custom stereo head rig & \cmark & \cmark & \cmark & \cmark & \cmark & \xmark & Home, retail, service \\
Xperience-10M~\cite{ropedia2026xperience} & 10,000 & 6-stream head rig + mocap & \cmark & \cmark & \cmark & \cmark & \cmark & \xmark & Daily life \\
EgoSuite-Open100K~\cite{lightwheel2026egosuite} & 10,000 & Head + wrist cameras & \xmark & \xmark & \xmark & \cmark & \xmark & \xmark & Home, retail, industry \\
Egocentric-10K~\cite{buildai2025egocentric10k} & 10,000 & Monocular head camera & \cmark & \xmark & \xmark & \xmark & \xmark & \cmark & Factories \\
\midrule
\textbf{RoboCap-10k (ours)} & \textbf{\nHours} & \textbf{RoboCap (6 cam, 2 IMU)} & \cmark & \cmark & \cmark & \cmark & \cmark & \cmark & \textbf{\nVerticals task families} \\
\bottomrule
\end{tabular*}
\caption{\textbf{Comparison with egocentric datasets used for manipulation learning.} A property is marked only if it holds for, and is reported across, the full released dataset. ``Occupational work'' is counted if the majority of the dataset is reported to be collected from individuals doing their occupational work. Hours count publicly released egocentric footage as of November 2026. RoboCap-10k is the only dataset combining calibrated stereo, metric head pose, and 3D hands with 10k+ released hours of occupational work.}
\label{tab:dataset_comparison}
\end{table*}

\section{Related Work}
\label{sec:related}

\subsection{Large-Scale Egocentric Datasets}

Early egocentric corpora~\cite{damen2018epickitchens,grauman2022ego4d} demonstrated the value of recording unscripted activity across thousands of hours and international cohorts, largely with head-mounted consumer cameras whose monocular, rolling-shutter footage does not support metric 3D reconstruction. Ego-Exo4D~\cite{grauman2024egoexo4d} and Nymeria~\cite{ma2024nymeria} moved to calibrated Aria glasses with external cameras or body motion capture, HOI4D~\cite{liu2022hoi4d} used RGB-D for category-level hand--object interaction, and EgoDex~\cite{hoque2025egodex} collected 829 hours of tabletop manipulation with on-device hand tracking from a mixed-reality headset.

A recent wave of datasets targets robot learning directly. Egocentric-10K~\cite{buildai2025egocentric10k} and its successors scale monocular factory footage to tens of thousands of hours or more, without 3D labels. EgoSuite-Open100K~\cite{lightwheel2026egosuite} has released the first 10,000 of a planned 100,000 hours of head- and wrist-camera footage across seven environment categories, including logistics and industry, with hand and body pose annotations and a license that permits commercial training. HumanNet~\cite{deng2026humannet} aggregates 1 million hours of human-centric video and recovers poses with monocular estimators. EgoVerse~\cite{punamiya2026egoverse} federates 1,362 hours from heterogeneous devices through a shared processing pipeline. EgoLive~\cite{li2026egolive} and Xperience-10M~\cite{ropedia2026xperience} use custom calibrated stereo rigs with hand and head annotations. RoboCap-10k differs from these in three respects (Table~\ref{tab:dataset_comparison}): all data comes from a single device model with two calibrated stereo baselines plus lateral coverage designed for the near field; calibration is re-estimated per session rather than assumed from the factory; and the content is occupational work across \nVerticals task families rather than daily-life or household routines.

Calibrated research devices~\cite{engel2023aria,lv2024ariaeveryday,pan2023adt} and multi-view hand--object benchmarks~\cite{banerjee2024hot3d,rim2026show3dcapturingscenes3d} provide accurate ground truth, but depend on expensive motion-capture rigs, markerless capture cages, or the built-in real-time trackers of consumer headsets~\cite{hoque2025egodex}, whose estimates cannot be re-run, inspected, or improved after capture, which confines them to tens of hours and small participant pools. Earlier wearable systems used downward-facing fisheye cameras on caps or body rigs for markerless body capture~\cite{rhodin2016egocap,xu2019mo2cap2,akada2022unrealego,akada20243dhumanposeperception}, but targeted the problem of body pose estimation rather than near-field hand manipulation.

Portable demonstration interfaces and hand-mounted rigs~\cite{chi2024umi,wang2024dexcap,tao2026dexwilddexteroushumaninteractions} record in-the-wild interactions with instrumented end-effectors or gloves, at the cost of altering natural hand use. The RoboCap leaves the hands bare.

\subsection{Egocentric Visual-Inertial SLAM}

Head-worn ego-motion estimation must cope with rapid rotation, close-range parallax, and the wearer's own hands occluding the view~\cite{krishnan2025benchmarkingegocentricvisualinertialslam,demayo2025monado}. Classical visual-inertial systems~\cite{qin2018vinsmono,leutenegger2015okvis,leutenegger2022okvis2,campos2021orbslam3,usenko2020basalt,geneva2019openvins,rosinol2020kimera} combine on-manifold IMU preintegration~\cite{forster2017preintegration,forster2017svo} with sliding-window optimization~\cite{demmel2021sqrt}. GPU implementations~\cite{korovko2025cuvslam} extend these to multi-camera rigs, learned dense pipelines~\cite{teed2021droidslam,murai2025mast3rslam,wang2025vggt} trade sparse optimization for dense correspondence, and Gaussian-splatting backends~\cite{zhu2026vigsslamvisualinertialgaussian,cao2026mcgsslammulticameraslamframework} improve offline mapping fidelity. All remain vulnerable when manipulation dominates the forward field of view.

Wearable rigs also deform. Offline toolkits~\cite{furgale2013kalibr} recover a rigid calibration at the factory, but mechanical and thermal stress in the field shift it. Across the RoboCap fleet, the relative rotation between stereo cameras drifts by 0.2--1.1$^\circ$ (median $\approx$0.5$^\circ$), producing 1--5\,px of epipolar residual and a 1--2\,cm depth bias at manipulation range~\cite{ali2026robocapnewplatformegocentric}. Continuous stereo self-calibration has been studied for vehicles~\cite{dang2009continuous}; we apply per-session rotational self-calibration to every recording in the corpus (Section~\ref{sec:ocal}).

\subsection{Egocentric Hand Pose Estimation}

Parametric hand models~\cite{romero2017embodied} and transformer-based monocular reconstructors~\cite{pavlakos2024hamer,potamias2024wilor,rong2021frankmocap,zhang2020mediapipehands} have made single-image 3D hand recovery robust in the wild, and world-space extensions recover trajectories from moving cameras~\cite{zhang2025hawor,yu2025dynhamr}. Monocular methods, however, regress depth under a weak-perspective assumption from apparent hand size, so their outputs are metrically ambiguous and inconsistent across views. Interaction benchmarks~\cite{chao2021dexycbbenchmarkcapturinghand,hampali2021ho3dv3improvingaccuracyhandobject,yang2022oakinklargescaleknowledgerepository} provide accurate multi-view supervision but are captured in controlled settings. Multi-view and learnable-triangulation methods~\cite{iskakov2019learnable,han2020megatrack,han2022umetrack,yang2023poem,poemv2} resolve the depth ambiguity by fusing synchronized views; production systems of this kind derive labels from motion capture. The Grounded API hand tracker is a feed-forward multi-view model trained on in-the-wild RoboCap recordings labeled mainly by offline multi-view optimization rather than motion capture~\cite{ali2026robocapnewplatformegocentric}.

\subsection{Egocentric Robot Learning}
Imitation learning from first-person human video~\cite{kareer2024egomimic,liu2025egozero} and cross-embodiment pretraining on human motion~\cite{qiu2025humanoidpolicy,ma2026humanscale,egoscale2026} show that human demonstrations transfer to robots once hand and head poses are mapped to a shared action space. Because retargeting consumes estimated poses, estimation error enters the policy's training distribution as label noise. RoboCap-10k is designed to reduce that noise at scale while keeping the raw multi-view streams, so labels can be regenerated as estimators improve.

\section{Preliminaries}
\label{sec:preliminaries}

We summarize the RoboCap hardware and Grounded API processing pipeline; full details and evaluations are in the companion report~\cite{ali2026robocapnewplatformegocentric}.

\subsection{The RoboCap}
\label{sec:robocap}

The RoboCap integrates its sensor payload into a 250\,g hat that distributes weight over the crown, avoiding the nose-bridge load of glasses during multi-hour sessions (Table~\ref{tab:sensor_specs}). Six global-shutter 1920$\times$1080 cameras run at 30\,Hz: a front and an eye stereo pair with fisheye optics give 157$^\circ$ and 154$^\circ$ of binocular overlap on 86.8\,mm and 106.2\,mm baselines, keeping the hands inside a metric stereo frustum across the reach envelope, and two lateral wide-angle cameras, mounted 85$^\circ$ off the forward axis and pitched 30$^\circ$ down, capture reaches that leave the stereo field and keep the trajectory solver constrained when the hands cover the front pair. All cameras are exposure-triggered by a single PWM signal fanned out 1-to-6, so frames are simultaneous by construction. Two 6-axis IMUs on opposite sides of the crown sample at 200\,Hz in the same 24\,MHz TCXO clock domain, with camera--IMU offsets identified per unit to a residual of at most 3\,ms. Streams are encoded on-device (H.265, 4\,Mbps per camera) to 214.5\,GB of onboard storage. A tethered 10,000\,mAh bank sustains roughly six hours of recording.

\begin{table}[t]
\centering
\footnotesize
\setlength{\tabcolsep}{3.5pt}
\begin{tabular}{@{}lccc@{}}
\toprule
\textbf{Subsystem} & \textbf{Optics / FoV} & \textbf{Baseline / Overlap} & \textbf{Rate / Sync} \\
\midrule
\makecell[l]{Front stereo} & \makecell{Fisheye \\ $180^\circ \times 112^\circ$} & \makecell{86.8\,mm \\ ($157^\circ$ overlap)} & \makecell{1080p @ 30\,Hz \\ HW trigger} \\
\addlinespace[0.3em]
\makecell[l]{Eye stereo}   & \makecell{Fisheye \\ $180^\circ \times 112^\circ$} & \makecell{106.2\,mm \\ ($154^\circ$ overlap)} & \makecell{1080p @ 30\,Hz \\ HW trigger} \\
\addlinespace[0.3em]
\makecell[l]{Lateral (2$\times$)} & \makecell{Wide-angle \\ $161^\circ \times 98^\circ$} & \makecell{$85^\circ$ off-axis, \\ $30^\circ$ down} & \makecell{1080p @ 30\,Hz \\ HW trigger} \\
\addlinespace[0.3em]
IMUs (2$\times$) & 6-axis, crown sides & Cam--IMU $\leq$3\,ms & 200\,Hz \\
\addlinespace[0.3em]
Clock & 24\,MHz TCXO & 1-to-6 PWM & $\pm$1.81\,ppm \\
\bottomrule
\end{tabular}
\caption{RoboCap sensor suite (adapted from~\cite{ali2026robocapnewplatformegocentric}). All cameras are global-shutter, 10-bit RGB with $f\!\approx\!590$--$630$\,px.}
\label{tab:sensor_specs}
\end{table}

\subsection{Factory and Online Calibration}
\label{sec:ocal}

Every unit is calibrated at the factory against an AprilGrid target with continuous-time batch optimization~\cite{furgale2013kalibr}, recovering pinhole intrinsics with 4-parameter Kannala--Brandt distortion, camera--camera and camera--IMU extrinsics, and time offsets. Units outside per-parameter tolerance gates are reworked before shipment.

Field drift is dominated by rotation~\cite{ali2026robocapnewplatformegocentric}. We therefore re-estimate the rotations of all six cameras jointly for each recording session, holding every factory translation fixed: epipolar constraints cannot observe baseline length, so metric scale stays anchored to the factory calibration. This extends the pairwise procedure of~\cite{ali2026robocapnewplatformegocentric}, which we revised after observing in-field deformation of deployed units. The left-front camera is the reference and is never moved; each other camera $c$ receives a rotation $\mathbf{R}_c \in SO(3)$ relative to it. For every camera pair $(a,b) \in \mathcal{P}$ with overlapping fields of view (the front and eye stereo pairs, the front--eye pairs, and each lateral camera with the stereo cameras it overlaps), correspondences $\mathcal{M}_{ab}$ are lifted to normalized image coordinates $\hat{\mathbf{x}} = \pi^{-1}(\mathbf{u})$ through the Kannala--Brandt model, and we solve
\begin{equation}
\min_{\{\mathbf{R}_c\}} \sum_{(a,b) \in \mathcal{P}} \sum_{i \in \mathcal{M}_{ab}} \rho_{\delta}\!\left( \frac{(\hat{\mathbf{x}}'^{\top}_{i} \mathbf{E}_{ab} \hat{\mathbf{x}}_{i})^2}{\|(\mathbf{E}_{ab}\hat{\mathbf{x}}_{i})_{1:2}\|^2 + \|(\mathbf{E}_{ab}^\top\hat{\mathbf{x}}'_{i})_{1:2}\|^2} \right),
\label{eq:calibration_objective}
\end{equation}
where $\mathbf{E}_{ab} = [\mathbf{t}^0_{ab}]_\times \mathbf{R}_b \mathbf{R}^{0}_{ab} \mathbf{R}_a^{\top}$ composes each camera's correction with the factory relative pose $(\mathbf{R}^0_{ab}, \mathbf{t}^0_{ab})$, and $\rho_\delta$ is a Huber kernel. Solving jointly keeps the corrections consistent across the rig: a camera that participates in several pairs receives a single rotation, so pairwise estimates cannot disagree. A session is rejected if any of the six cameras is degenerate, or if the feature-matching reprojection error after correction exceeds 1\,px on the front cameras or 1.5\,px on the side cameras.

Across the corpus, online calibration brings every device below 1\,px of average Sampson error, with a median device of 0.62\,px on the front stereo pair and 0.62\,px on the eye pair (Section~\ref{sec:exp_calib}).

\subsection{Visual-Inertial SLAM}
\label{sec:slam}

Head trajectories are estimated in a metric, gravity-aligned world frame by Grounded API's multi-camera visual-inertial SLAM~\cite{ali2026robocapnewplatformegocentric}. A GPU-accelerated patch-based optical-flow front end feeds a square-root sliding-window estimator~\cite{usenko2020basalt,demmel2021sqrt} with preintegrated inertial factors~\cite{forster2017preintegration}. On the RoboCap the front pair forms a stereo rig that gives landmarks metric depth from their first observation, while the two lateral cameras (four cameras in total; the eye pair is not used for SLAM) track their own landmarks monocularly, keeping the estimator constrained when hands and tools cover the front views. Appearance-based place recognition detects revisits, and the released trajectories come from an offline keyframe bundle adjustment over the whole session once all revisits are known, which is roughly twice as accurate as the causal estimate available during capture. One configuration is used for every session.

\begin{figure*}[t]
\centering
\includegraphics[width=\textwidth]{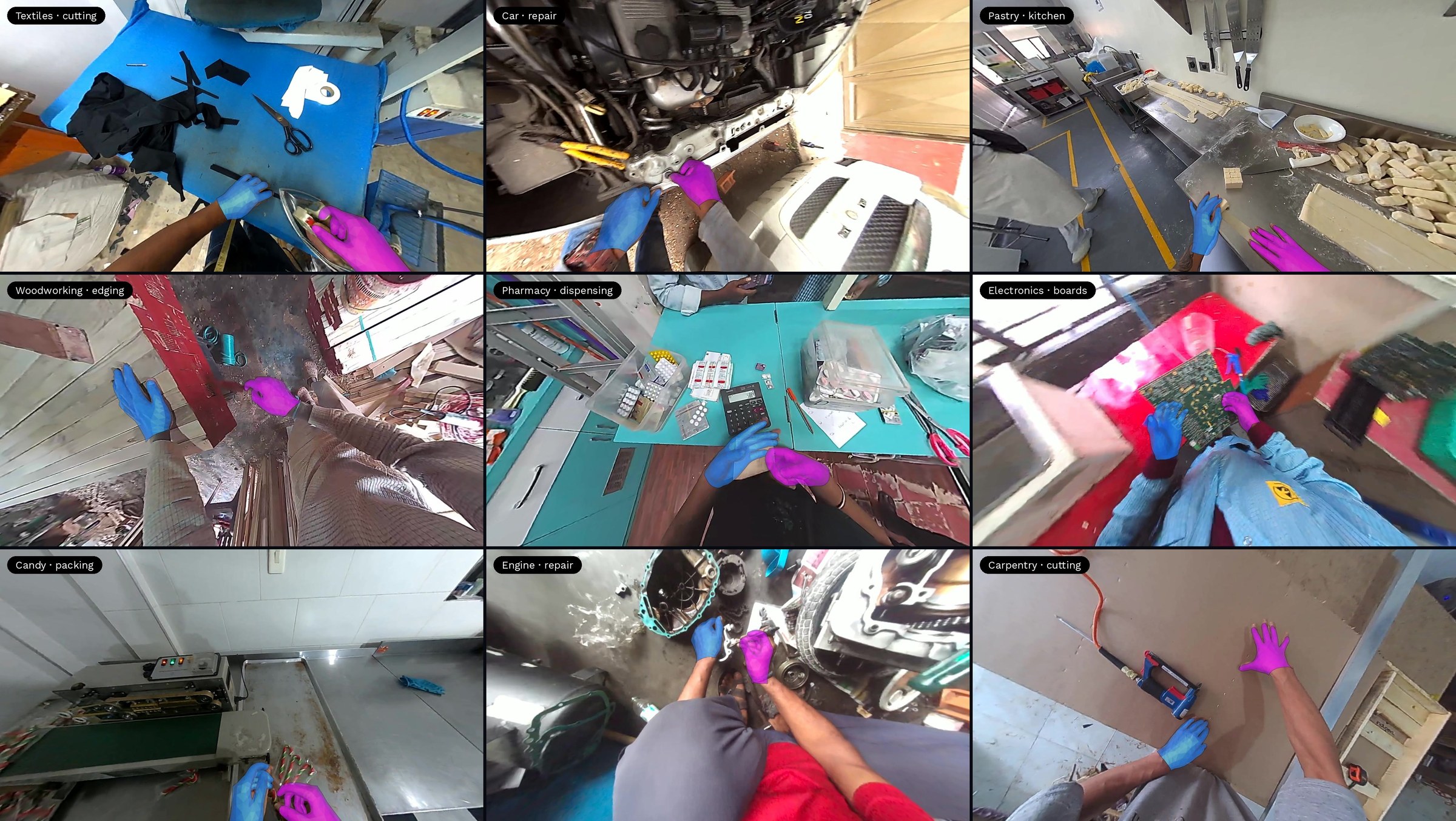}
\caption{\textbf{Hand pose estimates across workplaces.} Egocentric frames from the nine sessions of Figure~\ref{fig:teaser}, with the estimated MANO hand meshes rendered over the image (blue: left hand, magenta: right hand). The tracker holds up through gloves, tool and object contact, partial occlusion, clutter and outdoor lighting. Appendix~\ref{sec:supp_hand_gallery} shows 36 further sessions.}
\label{fig:ego_hands}
\end{figure*}

\subsection{Hand Pose Estimation}
\label{sec:hand}

RoboCap-10k provides per-frame metric hand poses for both hands as MANO~\cite{romero2017embodied} parameters together with 21 keypoints (16 joints plus 5 fingertips), expressed in the rig frame (left-front camera) and placed in the world frame by the head pose. The Grounded API tracker is a feed-forward multi-view ViT that takes hand crops from all cameras that see the hand, together with their calibrated intrinsics and extrinsics, and predicts a single MANO pose consistent across views by construction. It is trained on a large in-the-wild RoboCap corpus whose labels come from an offline multi-view optimization that bootstraps monocular priors~\cite{pavlakos2024hamer,potamias2024wilor} and refines a single MANO fit per trajectory for reprojected appearance and plausibility. The training data also includes a small motion-capture split recorded on the RoboCap (Section~\ref{sec:exp_hand}). This gives robustness to motion blur, tool occlusion, gloves, and difficult lighting. Figure~\ref{fig:ego_hands} shows the resulting hand meshes across workplaces, rendered from the released MANO parameters. Section~\ref{sec:exp_hand} summarizes its accuracy.

\begin{table*}[t]
\centering
\small
\setlength{\tabcolsep}{0pt}
\renewcommand{\arraystretch}{1.12}
\begin{tabular}{@{}p{5.2cm}*{6}{>{\centering\arraybackslash}p{1.95cm}}@{}}
\toprule
\textbf{Task family} & \textbf{Hours (\%)} & \textbf{Task types} & \textbf{Tasks} & \textbf{Easy} & \textbf{Medium} & \textbf{Hard} \\
\midrule
Machine Sewing \& Stitching & 14.9 & \cellcolor{figgreen!26}13 & \cellcolor{figgreen!38}191 & \cellcolor{figgreen!5}18 & \cellcolor{figgreen!18}47 & \cellcolor{figgreen!38}126 \\
Agricultural Harvesting \& Farming & 10.7 & \cellcolor{figgreen!30}15 & \cellcolor{figgreen!28}143 & \cellcolor{figgreen!28}95 & \cellcolor{figgreen!16}44 & \cellcolor{figgreen!4}4 \\
Food Cutting \& Processing & 9.6 & \cellcolor{figgreen!32}16 & \cellcolor{figgreen!37}188 & \cellcolor{figgreen!22}75 & \cellcolor{figgreen!38}102 & \cellcolor{figgreen!4}11 \\
Manual Crafting \& Assembly & 9.4 & \cellcolor{figgreen!32}16 & \cellcolor{figgreen!35}177 & \cellcolor{figgreen!13}44 & \cellcolor{figgreen!31}83 & \cellcolor{figgreen!15}50 \\
Packaging \& Material Bagging & 8.8 & \cellcolor{figgreen!30}15 & \cellcolor{figgreen!38}190 & \cellcolor{figgreen!36}123 & \cellcolor{figgreen!23}62 & \cellcolor{figgreen!4}5 \\
Textile \& Garment Finishing & 7.8 & \cellcolor{figgreen!28}14 & \cellcolor{figgreen!24}120 & \cellcolor{figgreen!19}63 & \cellcolor{figgreen!18}49 & \cellcolor{figgreen!4}8 \\
Cleaning \& Sanitation & 6.5 & \cellcolor{figgreen!38}19 & \cellcolor{figgreen!30}152 & \cellcolor{figgreen!38}129 & \cellcolor{figgreen!7}19 & \cellcolor{figgreen!4}4 \\
Metalworking \& Heavy Machining & 4.1 & \cellcolor{figgreen!22}11 & \cellcolor{figgreen!21}106 & \cellcolor{figgreen!4}13 & \cellcolor{figgreen!13}34 & \cellcolor{figgreen!18}59 \\
Woodworking \& Carpentry & 3.8 & \cellcolor{figgreen!20}10 & \cellcolor{figgreen!18}91 & \cellcolor{figgreen!4}8 & \cellcolor{figgreen!19}50 & \cellcolor{figgreen!10}33 \\
Retail Sales \& Inventory & 3.7 & \cellcolor{figgreen!30}15 & \cellcolor{figgreen!24}119 & \cellcolor{figgreen!16}55 & \cellcolor{figgreen!20}54 & \cellcolor{figgreen!4}10 \\
Digital Interaction \& Workplace Tasks & 3.4 & \cellcolor{figgreen!18}9 & \cellcolor{figgreen!17}84 & \cellcolor{figgreen!15}52 & \cellcolor{figgreen!11}30 & \cellcolor{figgreen!4}2 \\
Cooking \& Culinary Tasks & 3.4 & \cellcolor{figgreen!18}9 & \cellcolor{figgreen!21}105 & \cellcolor{figgreen!10}34 & \cellcolor{figgreen!25}68 & \cellcolor{figgreen!4}3 \\
Plastics \& Pipe Fabrication & 3.0 & \cellcolor{figgreen!12}6 & \cellcolor{figgreen!11}56 & \cellcolor{figgreen!9}32 & \cellcolor{figgreen!8}22 & \cellcolor{figgreen!4}2 \\
Furniture Assembly \& Upholstery & 3.0 & \cellcolor{figgreen!18}9 & \cellcolor{figgreen!14}69 & \cellcolor{figgreen!4}2 & \cellcolor{figgreen!12}33 & \cellcolor{figgreen!10}34 \\
Vehicle Maintenance \& Repair & 2.8 & \cellcolor{figgreen!14}7 & \cellcolor{figgreen!15}76 & \cellcolor{figgreen!4}7 & \cellcolor{figgreen!5}14 & \cellcolor{figgreen!17}55 \\
Personal Care \& Animal Services & 2.1 & \cellcolor{figgreen!8}4 & \cellcolor{figgreen!7}37 & \cellcolor{figgreen!4}8 & \cellcolor{figgreen!5}13 & \cellcolor{figgreen!5}16 \\
Electronics \& Circuit Assembly & 1.8 & \cellcolor{figgreen!10}5 & \cellcolor{figgreen!10}51 & \cellcolor{figgreen!4}6 & \cellcolor{figgreen!8}22 & \cellcolor{figgreen!7}23 \\
Building Construction \& Masonry & 1.3 & \cellcolor{figgreen!14}7 & \cellcolor{figgreen!9}45 & \cellcolor{figgreen!4}5 & \cellcolor{figgreen!6}15 & \cellcolor{figgreen!8}25 \\
\midrule
\textbf{Total} & \textbf{100.0} & \textbf{\nSiteTypes} & \textbf{\nTasks} & \textbf{\nTasksEasy} & \textbf{\nTasksMedium} & \textbf{\nTasksHard} \\
\bottomrule
\end{tabular}
\caption{\textbf{RoboCap-10k task taxonomy}, sorted by share of hours; shading of the count columns is proportional to each column's maximum. Tasks are clusters of the released task captions; task types group related tasks, and task families group related types (Section~\ref{sec:corpus_taxonomy}). Difficulty is rated per task.}
\label{tab:diversity_matrix}
\end{table*}

\section{The RoboCap-10k Corpus}
\label{sec:corpus}

\begin{figure*}[t]
\centering
\begin{subfigure}[t]{0.465\textwidth}
  \includegraphics[width=\linewidth]{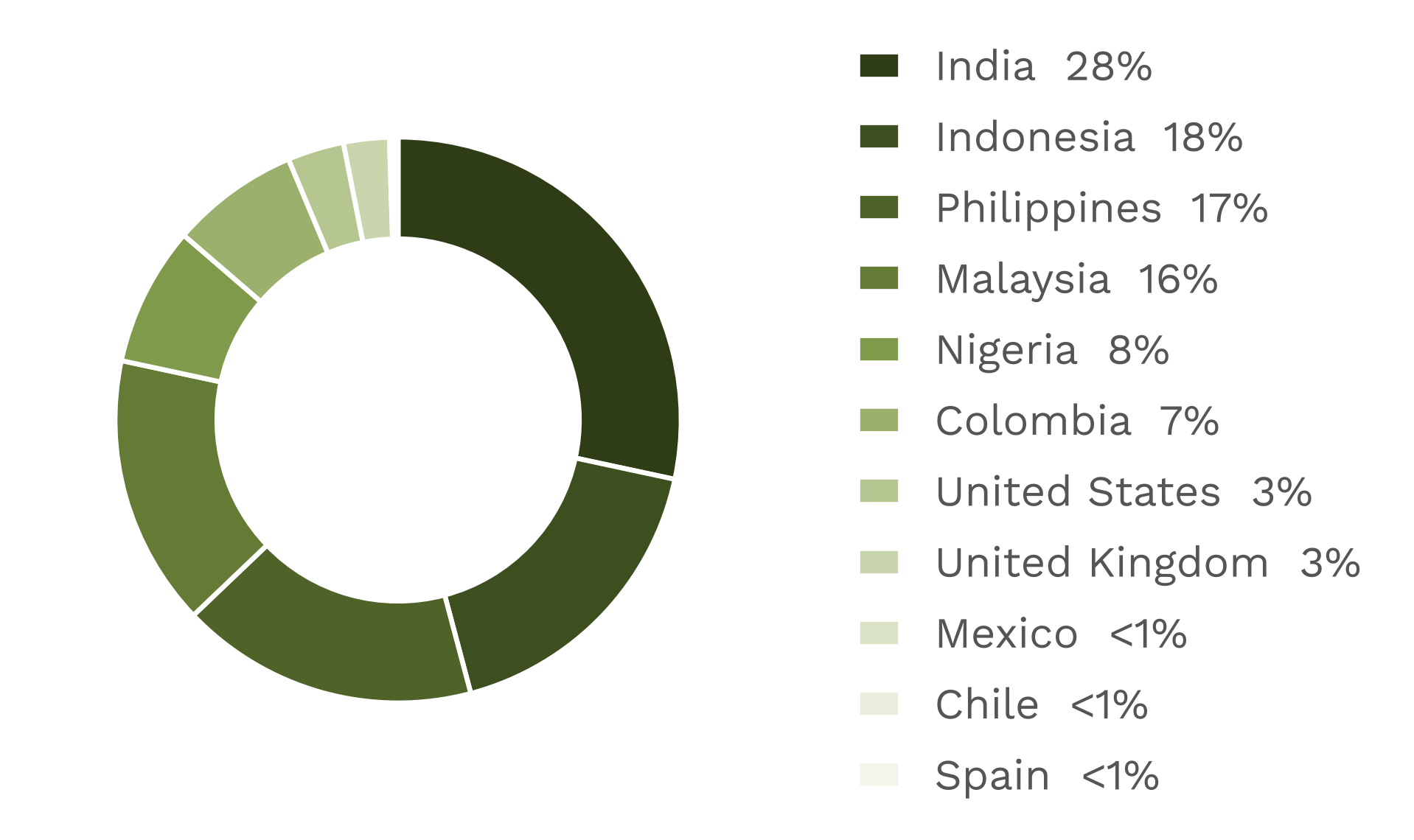}
  \caption{Released hours by country.}
  \label{fig:hours}
\end{subfigure}\hfill
\begin{subfigure}[t]{0.465\textwidth}
  \includegraphics[width=\linewidth]{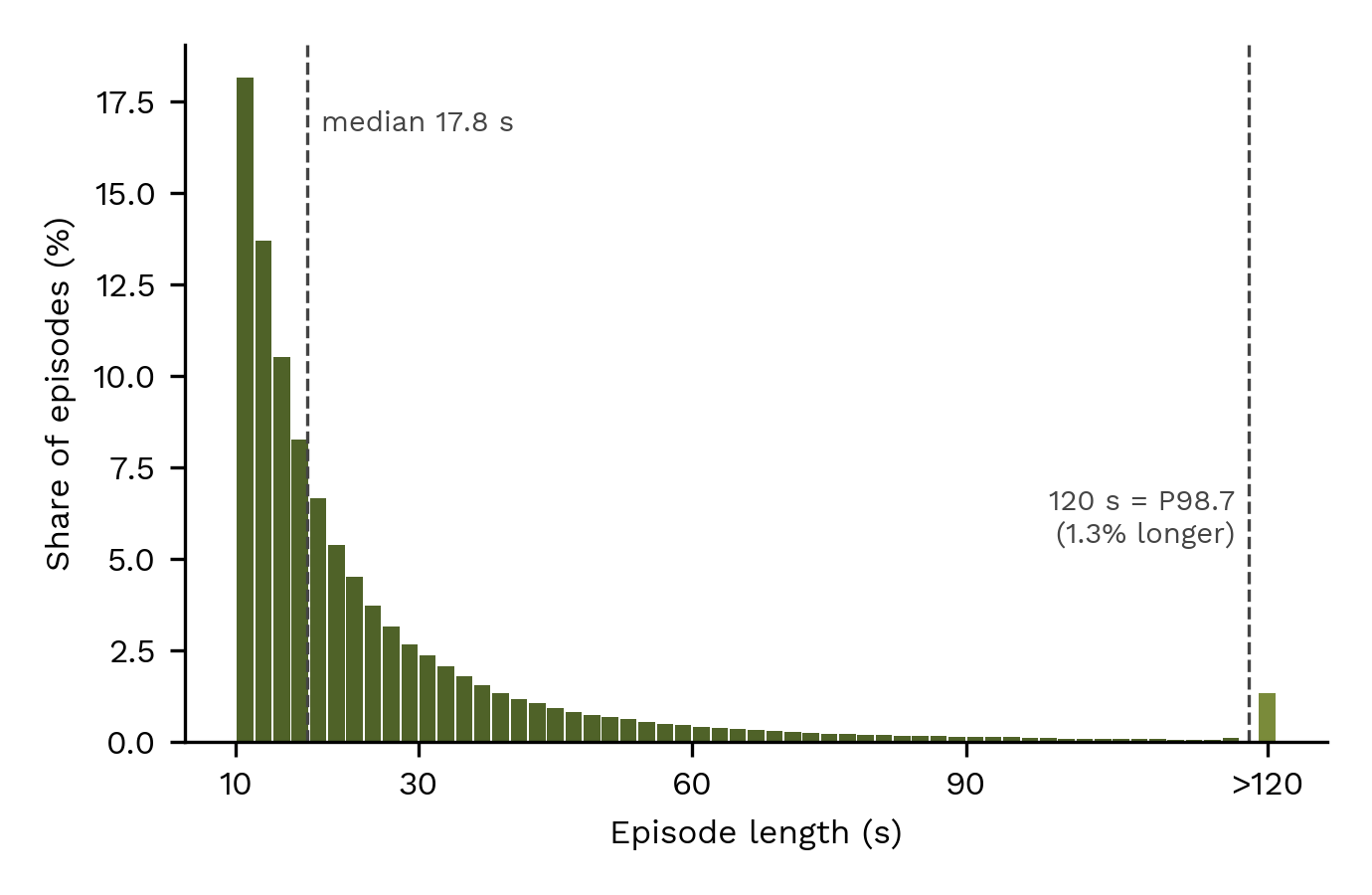}
  \caption{Episode length distribution.}
  \label{fig:episode_lengths}
\end{subfigure}
\caption{\textbf{Corpus statistics.} (a)~Share of released hours by country. (b)~Share of episodes by length in 2\,s bins; episodes longer than 120\,s are grouped in the final bar; dashed lines mark the median and the 120\,s cut-off, which falls at the 98.7th percentile.}
\label{fig:corpus_stats}

\vspace{2em}

\centering
\includegraphics[width=\textwidth]{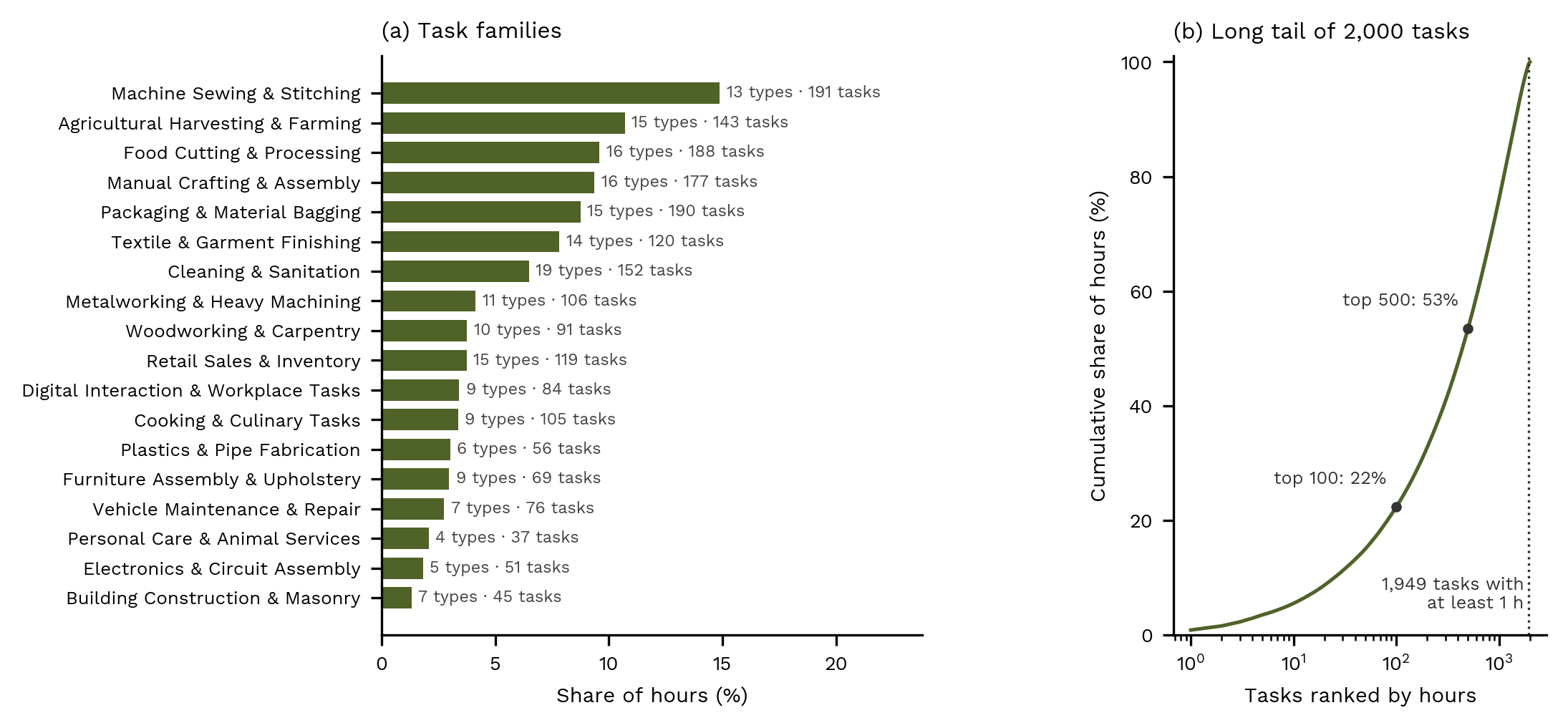}
\caption{\textbf{RoboCap-10k taxonomy}. (a)~Share of hours per task family, with its number of task types and tasks. (b)~The \nTasks tasks ranked by hours: cumulative share of footage. See Table~\ref{tab:diversity_matrix} for full details.}
\label{fig:tax_verticals}
\label{fig:taxonomy}
\end{figure*}

RoboCap-10k comprises \nHours hours (\nEpisodes episodes) of unscripted egocentric manipulation recorded during real work. An episode is one continuous clip of at least 10\,s in which both hands are tracked and SLAM is available; episodes run from 10\,s to 600\,s (median 17.8\,s, mean 26.1\,s, 90\% under 48.2\,s). These are curated hours, not raw recording time: each is retained only if it shows active bimanual manipulation with an identifiable task and passes the calibration, SLAM, and hand-tracking checks in Section~\ref{sec:experiments}. The corpus totals about \nTimesteps synchronized timesteps and \nFrames camera frames across the six streams. Hours are wall-clock capture time, and each hour contains six synchronized video streams. Figure~\ref{fig:hours} shows how the hours are distributed across countries.

\subsection{Design Principles}
\label{sec:principles}

We design RoboCap-10k around four principles.

\paragraph{Environment diversity.}
Policies degrade sharply under visual distribution shift. RoboCap-10k records activity in real workplaces, from bakeries and garment factories to car repair garages, PCB assembly lines, and clinical diagnostic labs, with their natural clutter, lighting, and tools left as they are.

\paragraph{Task diversity and hierarchy.}
Manipulation at work is bimanual, contact-rich, and long-horizon. The corpus covers \nTasks distinct tasks, including tool use, assembly, deformable-object handling, and fine electronics work. Each episode pairs an episode-level description of intent with dense sub-task segments, supervising both task planning and short-horizon motor control.

\paragraph{Metric action labels.}
Learning control from uncalibrated video leaves models to solve an ill-posed inverse problem. RoboCap-10k pairs every synchronized frame with a gravity-aligned 6-DoF head pose and metric MANO hand poses in the rig frame, which together define the world-frame 3D end-effector trajectories used to retarget human motion to robots.

\paragraph{Standardized hardware and calibration.}
All data comes from one device model~\cite{ali2026robocapnewplatformegocentric}, so imagery and 3D label quality are consistent across the corpus. Per-session online calibration removes the extrinsic drift that accumulates in deployed units, without targets or operator intervention.

\subsection{Collection Protocol}
\label{sec:corpus_collection}

Data was collected through partner enterprises in \nCountries countries. Wearers are existing employees who record during their regular paid shifts of 6--12\,h; the hat is the only change to their work, and tasks are not scripted, staged, or repeated for the camera. A local lead in each country recruits partner enterprises and coordinates the staff who rotate devices and train new wearers. Each unit shipped with its own factory calibration. Caps stay on site; every 1--3 days their recordings are offloaded to storage media, transported to a regional data center, and uploaded to the cloud, where Grounded API processes each session independently.

After processing, footage is curated into the released corpus at the level of episodes: contiguous segments in which one person performs one recognizable task in one setting. Trained annotators segment recordings into episodes following written guidelines that define task boundaries and exclude unusable footage (transit between work points, idle time, obstructed or unstable views), and every segmentation is reviewed by a second person; segmentations that fail review are returned for revision rather than silently corrected. A label is counted only if it was entered by an annotator, confirmed by a reviewer, or produced by a labeling stage validated against human annotation, and episodes missing any required label are held back. Labels are drawn from a centrally maintained, versioned taxonomy; automated checks flag implausible label combinations, and synonymous labels are merged. Each episode is further checked for complete, consistently timed camera streams and is deduplicated against all other episodes by file fingerprint, content signature, and recording metadata.

Every released episode is assigned a task, a task type and a task family from the derived taxonomy of Section~\ref{sec:corpus_taxonomy}, computed from its own captions, so the same procedure labels the entire corpus and can be re-run as the corpus grows.

\paragraph{Consent and ethics.}
Every recorded operator signed an informed consent form, and every partner enterprise signed a data agreement before recording began; participation was voluntary. Wearers are compensated on top of their regular wages and footage is never used to evaluate individual workers' performance. Compliance with local labor and privacy law is the responsibility of the partner operating in each country. No audio is captured, faces are anonymized in all released video (Section~\ref{sec:corpus_privacy}), and released metadata never identifies the site, employer, or participant.

\begin{table*}[t]
\centering
\footnotesize
\begin{tabularx}{\textwidth}{@{}l>{\raggedright\arraybackslash}X>{\raggedright\arraybackslash}X@{}}
\toprule
\textbf{Category} & \textbf{Fields} & \textbf{Description} \\
\midrule
\textbf{Episode} & \texttt{episode\_id}, \texttt{caption}, \texttt{scene}, \texttt{tasks}, \texttt{n\_frames}, \texttt{episode\_start\_ns}, \texttt{episode\_end\_ns}, \texttt{frame\_basis}, \texttt{time\_basis} & Episode identity, the episode-level caption (\texttt{scene} plus high-level task), the list of high-level tasks, and the frame and clock conventions shared by all units. \\
\addlinespace[0.4em]
\textbf{Unit timing} & \texttt{start\_frame}, \texttt{end\_frame}, \texttt{start\_s}, \texttt{end\_s}, \texttt{start\_ns}, \texttt{end\_ns}, \texttt{source\_start\_frame}, \texttt{source\_end\_frame} & Half-open boundaries in clip frames, clip-relative seconds, device-clock nanoseconds, and frames of the source recording. \\
\addlinespace[0.4em]
\textbf{Unit text} & \texttt{task\_id}, \texttt{high\_level\_task}, \texttt{low\_level\_task} & The sub-task the unit belongs to and the local physical action. \\
\addlinespace[0.4em]
\multicolumn{3}{@{}p{\textwidth}@{}}{} \\
\bottomrule
\end{tabularx}
\caption{\textbf{Fields stored with each episode's dense captions.} Stored as one JSON object per episode (\texttt{dense\_captions} in \texttt{meta/episodes}); each frame also carries a \texttt{task\_index} and \texttt{subtask\_index} into the high- and low-level task tables.}
\label{tab:unit_metadata}

\vspace{2em}

\centering
\footnotesize
\begin{tabularx}{\textwidth}{@{}l>{\raggedright\arraybackslash}X@{}p{1.5em}@{}l>{\raggedright\arraybackslash}X@{}}
\toprule
\multicolumn{2}{@{}l}{\textbf{One episode}: its sparse caption and three dense units} & &
\multicolumn{2}{@{}l@{}}{\textbf{One dense unit}: all fields of unit 3 (bold, left)} \\
\cmidrule(r){1-2}\cmidrule(l{0pt}){4-5}
Sparse, 0.00--14.80\,s & \textit{Tears a piece of paper towel from a roll, folds it, and wipes a glass beaker.} & &
Start / end & 9.50--14.80\,s (frames 285--444), clip-relative \\
Dense, 0.00--3.50\,s & \textit{Tears a section of paper towel from the roll.} & &
\texttt{scene} & \textit{A person is standing at a laboratory workbench handling paper towels and a small container.} \\
Dense, 3.50--9.50\,s & \textit{Folds the torn paper towel sheet.} & &
\texttt{high\_level\_task} & \textit{Cleans a small container using a folded paper towel.} \\
\textbf{Dense, 9.50--14.80\,s} & \textit{\textbf{Wipes the interior of a small clear container with the folded paper.}} & &
\texttt{low\_level\_task} & \textit{Wipes the interior of a small clear container with the folded paper.} \\
\bottomrule
\end{tabularx}
\caption{\textbf{The two caption tiers on one episode.} \emph{Left:} one sparse caption describes the whole episode, and dense units split it into consecutive segments; each segment is listed here by its \texttt{low\_level\_task}. \emph{Right:} the fields stored for one dense unit (unit 3, bold on the left): its time span, a \texttt{scene} description, the sub-task it belongs to (\texttt{high\_level\_task}), and the physical action (\texttt{low\_level\_task}). }
\label{tab:caption_example}
\end{table*}

\subsection{Dataset Taxonomy}
\label{sec:corpus_taxonomy}

RoboCap-10k is organized by a three-level task taxonomy derived from its own captions (Table~\ref{tab:diversity_matrix}). Every episode's task caption is embedded with a sentence encoder and clustered: \nTasks \emph{tasks} (fine clusters), grouped into \nSiteTypes \emph{task types}, which form \nVerticals \emph{task families}. Every episode belongs to exactly one task, type and family; no footage is left unassigned. Names for each task, type and family are generated from their most common captions, and the appendix lists every task type with example tasks (Table~\ref{tab:supp_taxonomy}). Difficulty is rated per task:
\begin{enumerate}
    \item \emph{Easy} tasks are single-step and need no training (\nTasksEasy tasks, e.g., sorting groundnuts, picking products from shelves).
    \item \emph{Medium} tasks are multi-step and need coordination or planning (\nTasksMedium, e.g., shaping dough into loaves, portioning meals into trays).
    \item \emph{Hard} tasks are skilled work requiring training, experience, or domain expertise, including manual crafts (\nTasksHard, e.g., soldering components onto PCB panels, tying rebar cages, weaving cane onto chair frames, stitching garments on industrial machines).
\end{enumerate}

The distribution is long-tailed (Figure~\ref{fig:tax_verticals}). The four largest families (Machine Sewing \& Stitching, Agricultural Harvesting \& Farming, Food Cutting \& Processing, and Manual Crafting \& Assembly) hold 45\% of hours and \nTasksTopFour of the \nTasks tasks; the 100 most common tasks cover 22\% of hours and the 500 most common 53\%, while 1{,}949 tasks have at least one hour of footage each.

\begin{figure*}[t]
\centering
\sbox{\facetablebox}{%
\begin{minipage}{\columnwidth}
\centering
\footnotesize
\setlength{\tabcolsep}{0pt}
\newcommand{\rp}[2]{#1\,{\color{black!45}/\,#2}}
\begin{tabular*}{\linewidth}{@{\extracolsep{\fill}}lcccc@{}}
\toprule
 & \multicolumn{2}{c}{\textbf{Ours}} & \multicolumn{2}{c}{\textbf{RetinaFace}} \\
\cmidrule(lr){2-3}\cmidrule(l){4-5}
False positives / frame & 0.2 & 0.05 & 0.2 & 0.05 \\
\midrule
R\,/\,P, IoU 0.3 & \rp{87.7}{77.4} & \rp{77.3}{92.4} & \rp{29.8}{54.0} & \rp{25.1}{79.6} \\
R\,/\,P, IoU 0.5 & \rp{82.0}{71.6} & \rp{73.4}{87.4} & \rp{28.5}{51.7} & \rp{24.6}{78.1} \\
\midrule
\multicolumn{5}{@{}l}{\textit{Recall on the hardest cases}} \\
\quad Small faces (24--32\,px) & 69.8 & 51.1 & 6.5 & 1.4 \\
\quad Eye cameras & 88.8 & 78.9 & 2.5 & 1.7 \\
\quad Image periphery & 85.1 & 71.2 & 18.1 & 15.8 \\
\bottomrule
\end{tabular*}
\end{minipage}}%
\begin{minipage}[t]{\columnwidth}
  \vspace{0pt}\centering
  \includegraphics[height=\dimexpr\ht\facetablebox+\dp\facetablebox\relax,
                   width=\linewidth,keepaspectratio]{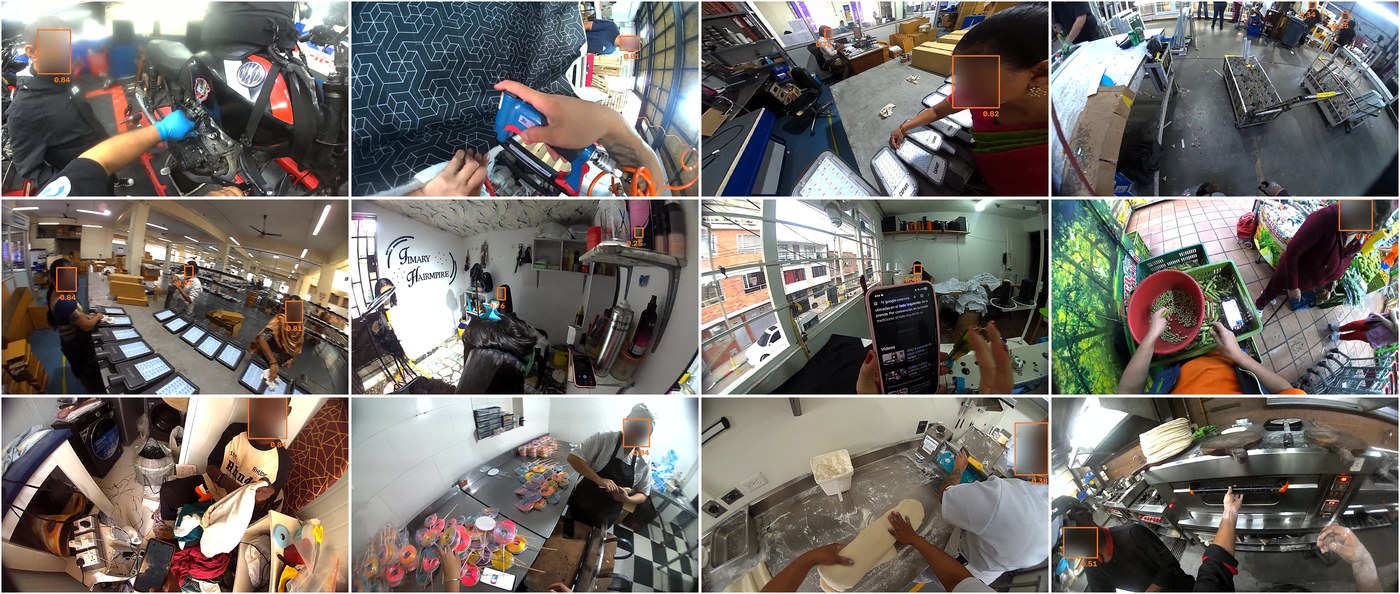}
  \captionof{figure}{\textbf{Face anonymization example} (from a front camera). Our face blurring model identifies faces across varied environments, demographics, camera models, and orientations in camera frame. See Figure~\ref{fig:supp_face_examples} for more examples.}
  \label{fig:face_example}
\end{minipage}\hfill
\begin{minipage}[t]{\columnwidth}
  \vspace{0pt}
  \usebox{\facetablebox}
  \captionof{table}{\textbf{Face detection on held-out sessions} (1{,}336 faces). Detectors are compared at thresholds giving 0.2 and 0.05 false positives per frame. Hardest-case rows report recall at IoU 0.3 for the smallest faces, the eye cameras, and the outer quarter of the image.}
  \label{tab:face_detection}
\end{minipage}
\end{figure*}

\subsection{Sparse and Dense Language Captions}
\label{sec:corpus_captioning}

To support language-conditioned policies and world models, RoboCap-10k provides two tiers of language. Rather than slot-filling a fixed taxonomy of objects and verbs, the captions describe what happens in free text and link fine-grained action segments to episode-level intent. We use two terms throughout. Each episode has exactly one \emph{sparse caption}: a single free-text sentence describing the whole clip. Each episode is also cut into consecutive \emph{dense units}: shorter time segments, each with its own structured caption made of the text fields below. Table~\ref{tab:caption_example} shows one episode with both tiers.

\paragraph{Tier 1: sparse captions (one per episode).}
Each episode carries one high-level caption of the overall task and object interaction, formed from the scene description and high-level task (e.g., \textit{``Tears a piece of paper towel from a roll, folds it, and wipes a glass beaker''} over 0.00--14.80\,s). This caption expresses the global intent of the clip and serves as the anchor for dense annotations.

\paragraph{Tier 2: dense units (many per episode).}
Each episode is temporally segmented into non-overlapping dense units, each covering one sub-task or action, with clip-relative start and end times in seconds and frames. A dense unit is not a separate caption type; it is a time segment, and its caption is spread across three text fields:
\begin{itemize}
    \item \texttt{scene}: the workspace and setting (e.g., \textit{``A person is standing at a laboratory workbench handling paper towels and a small container''});
    \item \texttt{high\_level\_task}: the sub-task being performed (e.g., \textit{``Cleans a small container using a folded paper towel''});
    \item \texttt{low\_level\_task}: the local physical action (e.g., \textit{``Wipes the interior of a small clear container with the folded paper''}).
\end{itemize}
 Dense units are generated by Gemini 3.8 Flash from the two front-stereo views, which keep the near field in view. The same model call produces the episode's \texttt{scene} and high-level tasks; the released episode-level caption is the \texttt{scene} followed by the high-level task, so the two tiers are consistent by construction rather than independent. Table~\ref{tab:unit_metadata} lists the stored fields. In short: the sparse caption answers \emph{what is this clip about}, and the dense units answer \emph{what is happening at each moment}, at two levels of detail (\texttt{high\_level\_task} for the sub-task, \texttt{low\_level\_task} for the physical action).

\subsection{Format and Release}
\label{sec:corpus_format}

RoboCap-10k is released under the Creative Commons Attribution 4.0 International license (CC BY 4.0). 
Every episode is provided in two formats. The LeRobot v3.0 format~\cite{cadene2024lerobot} is intended for training and is hosted on Hugging Face (\texttt{groundedsi/RoboCap-10k}, with a project page at \href{https://robocap10k.com}{\texttt{robocap10k.com}}) as 412 self-contained shards of about 25\,h each, $\sim$127\,TB in total. Each of the six cameras is stored as an undistorted, face-blurred H.265 MP4, and per-frame data is stored in Parquet tables indexed by frame: the SLAM head pose, and MANO parameters and 21 keypoints for both hands, expressed in the rig (left-front camera) frame and mapped to the world frame by the head pose, with per-hand validity flags. Per-episode metadata holds the captions (Table~\ref{tab:unit_metadata}), and each shard's \texttt{meta/cameras.json} holds every episode's intrinsics and extrinsics. The corpus plugs directly into existing LeRobot dataloaders. A Rerun recording (\texttt{.rrd}) of the same episode, hosted by Rerun, is intended for inspection: it places the six camera streams, the head trajectory, the hand keypoints, and the dense captions in a shared 3D world frame on a common timeline, and carries the MANO parameters as data. All streams share one device clock, so video frames, poses, and captions correspond one-to-one by frame index.

\begin{figure*}[t]
\centering
\sbox{\ocaltablebox}{%
\begin{minipage}{\columnwidth}
\centering
\footnotesize
\setlength{\tabcolsep}{0pt}
\begin{tabular*}{\linewidth}{@{\extracolsep{\fill}}lcc@{}}
\toprule
 & Front stereo (px) & Eye stereo (px) \\
\midrule
Median & 0.62 & 0.62 \\
Mean $\pm$ $\sigma$ & 0.64 $\pm$ 0.11 & 0.63 $\pm$ 0.11 \\
P5--P95 & 0.48--0.85 & 0.47--0.83 \\
Best device & 0.33 & 0.33 \\
Worst device & 0.99 & $<$1.00 \\
\bottomrule
\end{tabular*}
\end{minipage}}%
\begin{minipage}[t]{\columnwidth}
  \vspace{0pt}\centering
  \includegraphics[width=\linewidth,
                   height=\dimexpr\ht\ocaltablebox+\dp\ocaltablebox\relax,
                   keepaspectratio]{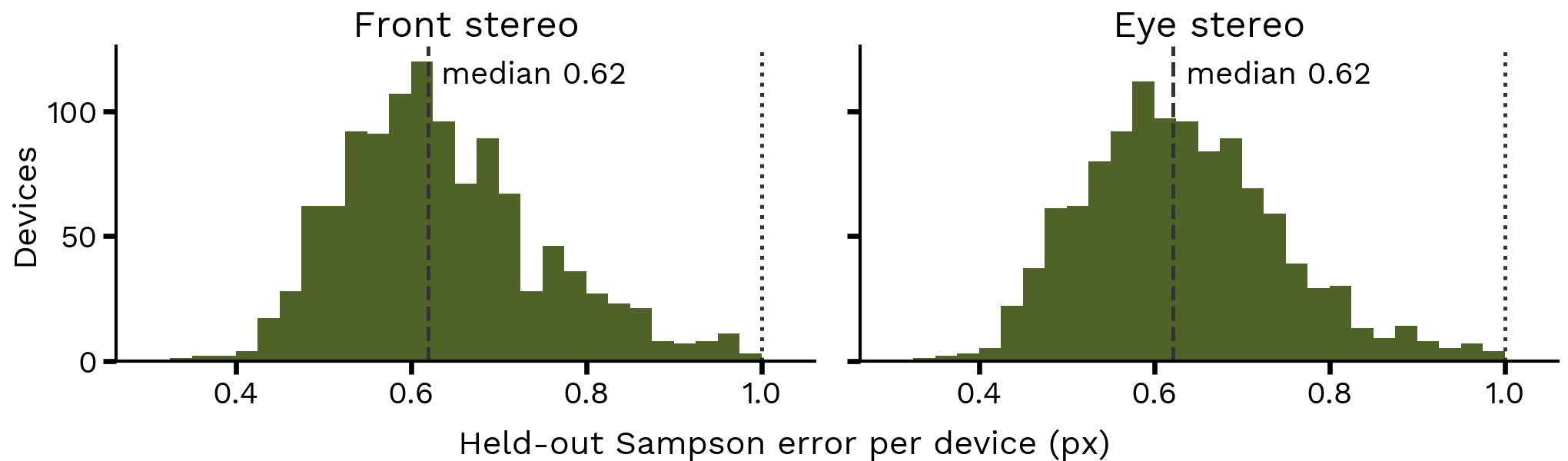}
  \captionof{figure}{\textbf{Calibration error per device after online calibration.} Held-out Sampson error of each device in the release. Dashed lines mark the median; dotted lines mark 1\,px.}
  \label{fig:ocal_spread}
\end{minipage}\hfill
\begin{minipage}[t]{\columnwidth}
  \vspace{0pt}
  \usebox{\ocaltablebox}
  \captionof{table}{\textbf{Distribution of per-device calibration error} (px) after online calibration, across the entire corpus. The spread is quasi-normal, slightly skewed towards lower errors.}
  \label{tab:ocal_spread}
\end{minipage}

\vspace{2em}

\centering
\footnotesize
\begin{minipage}[t]{0.48\textwidth}
  \centering
  \begin{tabular*}{\linewidth}{@{\extracolsep{\fill}}cccc@{}}
    \toprule
    Trajectories & Duration (h) & Median ATE (mm) & Mean ATE (mm) \\
    \midrule
    4,669 & 153.4 & 3.0 & 3.19 \\
    \bottomrule
  \end{tabular*}
  \captionof{table}{\textbf{Head trajectory accuracy} against motion capture on RoboMoCap~\cite{deshmukh2026robomocaplargescalewholebodymotion}. We compute ATE after $\mathrm{SE}(3)$ alignment.}
  \label{tab:slam_mocap}
\end{minipage}\hfill
\begin{minipage}[t]{0.48\textwidth}
  \centering
  \begin{tabular*}{\linewidth}{@{\extracolsep{\fill}}lcc@{}}
    \toprule
    Benchmark & Metric & Error (mm) \\
    \midrule
    RoboMoCap test~\cite{deshmukh2026robomocaplargescalewholebodymotion} & PA-MPJPE & 4.5 \\
    GSI-Internal~\cite{ali2026robocapnewplatformegocentric} & MPJPE & 9.2 \\
    \bottomrule
  \end{tabular*}
  \captionof{table}{\textbf{Hand reconstruction accuracy} over 21 keypoints. RoboMoCap: 9.2 min of held-out motion capture across 4 sessions, fitted to MANO. GSI-Internal: simulated and motion-capture sequences with known hand shape, reported in~\cite{ali2026robocapnewplatformegocentric}.}
  \label{tab:hand_mocap}
\end{minipage}
\end{figure*}

\subsection{Face Anonymization}
\label{sec:corpus_privacy}

Egocentric workplace footage routinely contains customers, co-workers, passersby, and occasionally the wearer in mirrors or reflective surfaces. Off-the-shelf face detectors are trained on eye-level photographs and miss many faces in this setting. For example, faces are seen from above through fisheye optics at the image periphery where distortion is strongest. Faces can also be small and distant, partially occluded by tools and hands, or blurred by head motion. We therefore train our own detector on roughly 1,000 hours of RoboCap footage with human-labeled face boxes, sampled across the collection countries, every camera position, and both indoor and outdoor sites.

The detector runs on every frame of all six streams. Its operating point balances two failure modes: a missed face is a privacy failure, while a false positive blurs part of the scene, and in egocentric footage that part is often the hands or the object being manipulated, which is exactly the signal the dataset exists to provide. We therefore release at 0.05 false positives per frame (Table~\ref{tab:face_detection}), where precision is 92.4\%. Detections are linked into tracks, propagated forward and backward along each track, and dilated before blurring, so a face missed in one frame is still covered by its neighbors. The dilation is 15\% of box size, kept small so that blur does not spread onto nearby hands. Faces are blurred with a Gaussian kernel, and the blur is irreversible: released videos are re-encoded from the blurred frames and the originals are not distributed.

\section{Annotation Quality}
\label{sec:experiments}

Absolute accuracy of Grounded API is established in the companion report against external benchmarks~\cite{ali2026robocapnewplatformegocentric}; we summarize it here and then report statistics computed over the corpus itself.

\subsection{Natural Language Captions}
Captions for 1{,}000 hours (10\% of the corpus) were visually reviewed by human annotators against the source video. We used these reviewed labels to tune the captioning pipeline, which then captioned the remaining 9{,}000 hours so that captions follow a standard ontology. Operators also provide basic metadata when they upload data to our services, including geography and occupation, which is used to further ground our captioning pipeline.

\subsection{Calibration Across the Corpus}
\label{sec:exp_calib}

We summarize calibration quality per device. For each device in the release, we average the held-out Sampson error after online calibration over all of its sessions, weighted by episode; the error of each camera includes all of its overlapping pairs, not only its stereo partner. Figure~\ref{fig:ocal_spread} and Table~\ref{tab:ocal_spread} show the resulting distribution across devices.

On the front stereo pair the median device error is 0.62\,px (mean 0.64\,px, standard deviation 0.11\,px), and the eye pair is similar (median 0.62\,px). Every device averages below 1\,px on both pairs, with the worst at 0.99\,px on the front pair and just under 1\,px on the eye pair.

\subsection{SLAM}
\label{sec:exp_slam}

The LaMAria benchmark~\cite{krishnan2025benchmarkingegocentricvisualinertialslam} is a head-worn visual-inertial benchmark of 63 city-scale test walks (27.5\,h) scored against surveyed control points. GroundedSLAM, Grounded API's SLAM implementation, achieves an average control-point score of 72.4, nearly matching the performance of Aria's SLAM, which scores 75.9.

We further evaluate trajectory accuracy directly against optical motion capture on RoboMoCap~\cite{deshmukh2026robomocaplargescalewholebodymotion}, where a NOKOV system tracks a rigid body on the RoboCap at 120\,Hz. We evaluate on 153.4 hours of motion capture data spread over the four headset units, recording dates, and seven activity families (locomotion, posture dynamics, loco-manipulation, whole-body environment interaction, dynamic recovery, static manipulation and long-horizon composites).

Each SLAM trajectory is metric and is evaluated as produced: the mocap clock offset is estimated from angular-speed correlation, the rigid-body-to-camera transform is solved per session with scale fixed to one, and the trajectory is aligned to the mocap with an $\mathrm{SE}(3)$ fit before computing the absolute trajectory error (ATE, position RMSE).

Table~\ref{tab:slam_mocap} summarizes the results.
On held-out sessions the median ATE is 3.0\,mm and the mean is 3.19\,mm. The metric scale is recovered to a median 0.17\,\% without any alignment scale, and the residual error is dominated by slow ($>$5\,s) drift rather than frame-to-frame noise, whose high-frequency component is about 0.2\,mm.

\subsection{Hand Tracking}
\label{sec:exp_hand}

We evaluate hand poses against motion capture on RoboMoCap~\cite{deshmukh2026robomocaplargescalewholebodymotion}, concurrent work recorded on the RoboCap. RoboMoCap's training split (34.9 minutes) is part of the tracker's training data, its validation split is used for checkpoint selection, and we report results on its held-out test split (9.2 minutes). RoboMoCap does not provide the operator's hand shape, so absolute joint error (MPJPE) would mix hand-size mismatch with tracking error. We therefore report Procrustes-aligned mean per-joint position error (PA-MPJPE) over the 21 hand keypoints, which removes global rotation, translation and scale and isolates finger articulation.

On the test split, Grounded API reaches a PA-MPJPE of 4.5\,mm (Table~\ref{tab:hand_mocap}), compared with 8.0\,mm for a model trained without the RoboMoCap split. Because the test split contains only held-out sequences, this gain reflects learning RoboMoCap's joint and wrist conventions rather than fitting frame-level label noise, so most of the 8.0\,mm is a convention mismatch rather than tracking error. For absolute error, we refer to the GSI-Internal benchmark of the companion report, a mix of simulated and motion-capture sequences where the hand shape is known, on which Grounded API reaches an MPJPE of 9.2\,mm~\cite{ali2026robocapnewplatformegocentric}.

\section{Limitations}
\label{sec:limitations}

\paragraph{Labels are estimates.}
All 3D annotations are produced by Grounded API, not measured. Calibration, SLAM and hand tracking are validated on external and RoboCap-specific benchmarks (Section~\ref{sec:experiments}), but millimeter ground truth cannot be collected in the wild, and hand errors concentrate where they matter most: fingers occluded at the contact interface during tight grasps. Language captions are model-generated; human review (about 1{,}000 hours) was used to tune the captioning pipeline, not to correct released captions.

\paragraph{Coverage and selection.}
Hours are spread over \nCountries countries, but the distribution is concentrated: India and the Philippines account for about two thirds of hours, while six countries contribute under 1\% each.  Curation keeps only bimanual manipulation that passes the 3D quality checks, so one-handed work, idle and transitional time, and the hardest tracking conditions (heavy occlusion, fast motion, poor lighting) are underrepresented relative to raw footage. Household and personal activities are largely absent, and task counts are uneven across task families.

\paragraph{No touch or force.}
There are no tactile, shear or wrench measurements, so downstream policies must infer contact from vision alone.

\paragraph{Embodiment gap.}
Human hands have far more degrees of freedom than most robot end-effectors, and the human arms visible in the images differ from robot arms. Retargeting to grippers or dexterous hands must still handle joint limits, singularities and this visual gap.

\section{Conclusion}
\label{sec:conclusion}

RoboCap-10k pairs the scale of egocentric video with metric 3D labels by standardizing on one purpose-built device, re-estimating its calibration for every session, and processing every hour with the same SLAM and multi-view hand-tracking stack. The result is \nHours hours of real occupational manipulation across \nCountries countries, \nVerticals task families and \nTasks distinct tasks, with world-frame head and hand trajectories and two tiers of language. We release it, with its raw multi-view streams, as pretraining data for generalist manipulation policies and world models, and as a testbed for a question the field has not yet answered: how much does the precision of human action labels matter at scale?

{
    \small
    \bibliographystyle{ieeenat_fullname}
    \bibliography{main}
}

\clearpage
\onecolumn
\setcounter{page}{1}
\begin{center}
  {\Large\textbf{\thetitle}}\\[0.5em]
  {\Large Supplementary Material}
\end{center}
\vspace{1em}

\appendix

\section{Task Taxonomy}
\label{sec:supp_taxonomy}
{\scriptsize
\renewcommand{\arraystretch}{1.1}
\setlength{\LTcapwidth}{\textwidth}
\begin{longtable}{@{}>{\raggedright\arraybackslash}p{3.8cm}>{\raggedright\arraybackslash}p{3.6cm}>{\raggedright\arraybackslash}p{7.4cm}>{\raggedright\arraybackslash}p{1.3cm}@{}}
\caption{\textbf{Representative task types}: the largest task type of each family, with two of its largest tasks and its typical difficulty. Table~\ref{tab:supp_taxonomy} lists all \nSiteTypes task types.}\label{tab:supp_site_types}\\
\toprule
\textbf{Task family} & \textbf{Task type} & \textbf{Example tasks} & \textbf{Difficulty} \\
\midrule
\endfirsthead
\toprule
\textbf{Task family} & \textbf{Task type} & \textbf{Example tasks} & \textbf{Difficulty} \\
\midrule
\endhead
\bottomrule
\endlastfoot
Machine Sewing and Stitching & Industrial machine garment sewing & Sewing fabric on sewing machine; Sewing fabric on a machine & Hard \\
Agricultural Harvesting and Farming & Tray sorting peanuts and seeds & Sorting peanuts on a tray; Sorting legumes on a tray & Easy \\
Food Cutting and Processing & Food assembly and plating & Prepping food ingredients on counter; Rolling and wrapping food wraps & Medium \\
Manual Crafting and Assembly & Textile weaving and crocheting & Crocheting yarn with a hook; Braiding natural fibers into rope & Hard \\
Packaging and Material Bagging & Small parts bagging and sealing & Bagging small parts; Packing items into plastic packaging & Easy \\
Textile and Garment Finishing & Cutting fabric with scissors & Cutting fabric with scissors; Trimming fabric pieces with scissors & Medium \\
Cleaning and Sanitation & Sink and basin dishwashing & Washing dishes in the sink; Washing dishes by hand & Easy \\
Metalworking and Heavy Machining & Welding and metal assembly & Welding metal structures; Assembling hardware components & Hard \\
Woodworking and Carpentry & Wood sawing and cutting & Trimming and sharpening thin sticks; Sawing wood with a handsaw & Hard \\
Retail Sales and Inventory & Produce weighing and bagging & Packing vegetables into plastic bags; Weighing and packaging produce & Medium \\
Digital Interaction and Workplace Tasks & Workplace smartphone operation & Operating a smartphone; Using a smartphone & Easy \\
Cooking and Culinary Tasks & Skewering grilling and deep frying & Skewering marinated food pieces; Stir-frying ingredients in a wok & Medium \\
Plastics and Pipe Fabrication & Handsaw plastic pipe cutting & Cutting plastic pipes with saw; Cutting plastic pipes with handsaw & Easy \\
Furniture Assembly and Upholstery & Cushion and chair frame upholstery & Upholstering furniture cushions; Upholstering furniture frame & Hard \\
Vehicle Maintenance and Repair & Motorcycle engine and component maintenance & Servicing motorcycle engine components; Cleaning motorcycle engine parts & Hard \\
Personal Care and Animal Services & Barbering and hair cutting & Cutting client hair with clippers; Cutting hair with scissors and clippers & Hard \\
Electronics and Circuit Assembly & Circuit board soldering and assembly & Soldering components onto circuit board; Assembling and inspecting circuit boards & Hard \\
Building Construction and Masonry & Masonry plastering and tiling & Applying mortar to masonry wall; Applying and smoothing wall plaster & Hard \\
\end{longtable}
}

\footnotesize
\setlength{\LTcapwidth}{\textwidth}
\renewcommand{\arraystretch}{1.05}
\begin{longtable}{@{}>{\raggedright\arraybackslash}p{3.7cm}>{\raggedright\arraybackslash}p{10.6cm}>{\centering\arraybackslash}p{0.9cm}>{\raggedright\arraybackslash}p{1.4cm}@{}}
\caption{\textbf{Complete task-type catalogue.} Every task type in RoboCap-10k, grouped by task family, with its largest tasks, its number of tasks and its typical difficulty.}
\label{tab:supp_taxonomy}\\
\toprule
\textbf{Task type} & \textbf{Example tasks} & \textbf{Tasks} & \textbf{Difficulty} \\
\midrule
\endfirsthead
\toprule
\textbf{Task type} & \textbf{Example tasks} & \textbf{Tasks} & \textbf{Difficulty} \\
\midrule
\endhead
\midrule
\multicolumn{4}{r@{}}{\textit{Continued on next page}} \\
\endfoot
\bottomrule
\endlastfoot
\multicolumn{4}{@{}l}{\textbf{Machine Sewing and Stitching} \textcolor{black!50}{(13 task types, 191 tasks)}} \\*
\cmidrule{1-4}\nopagebreak
Industrial machine garment sewing & Sewing fabric on sewing machine; Sewing fabric on a machine; Sewing fabric with sewing machine & 36 & Hard \\
Operating knitting and sewing machines & Stitching fabric on a sewing machine; Operating a manual knitting machine; Setting up sewing machine workstation & 27 & Hard \\
Sewing dark panels and leather & Stitching fabric on sewing machine; Sewing fabric panels on machine; Sewing dark fabric with machine & 23 & Hard \\
Hand sewing and craft assembly & Preparing and stitching fabric pieces; Sewing garment pieces by hand; Ironing and trimming fabric strips & 24 & Medium \\
Overlock machine sewing & Stitching fabric on overlock machine; Operating overlock sewing machine; Sewing fabric with overlock machine & 12 & Hard \\
Sewing blue garments and tape & Sewing fabric on a machine; Sewing garments and reflective tape; Aligning and sewing blue fabric & 15 & Hard \\
Sewing workstation with phone use & Using smartphone at workstation; Sewing garments with machine; Trimming loose garment threads & 16 & Easy \\
Red garment stitching and trimming & Stitching red fabric on machine; Trimming loose garment threads; Inspecting and trimming garment threads & 7 & Hard \\
Sewing and crafting pink garments & Overlocking pink fabric pieces; Trimming loose threads from garments; Crafting and sewing pink items & 6 & Hard \\
Preparing and trimming sewing fabric & Preparing fabric at sewing station; Trimming fabric at sewing table; Machine sewing fabric pieces & 9 & Hard \\
Sewing black fabric garments & Sewing black fabric garments; Machine sewing black fabric panels; Sewing fabric with overlock machine & 8 & Hard \\
Sewing green garments and packing & Sewing garments on a machine; Sewing and crocheting green fabric; Inspecting and trimming garments & 7 & Medium \\
Electric fabric cutting & Cutting fabric with electric cutter & 1 & Hard \\
\addlinespace[0.5em]
\multicolumn{4}{@{}l}{\textbf{Agricultural Harvesting and Farming} \textcolor{black!50}{(15 task types, 143 tasks)}} \\*
\cmidrule{1-4}\nopagebreak
Tray sorting peanuts and seeds & Sorting peanuts on a tray; Sorting legumes on a tray; Sorting raw peanuts by hand & 31 & Easy \\
Harvesting and trimming onions & Harvesting and processing onions; Harvesting and trimming field onions; Trimming and sorting harvested onions & 17 & Easy \\
Sorting and bundling vegetables & Sorting and bunching leafy greens; Sorting and bundling leafy greens; Bundling leafy green vegetables & 12 & Easy \\
Stripping leaves and stalks & Stripping leaves from plant stalks; Plucking leaves from stems; Stripping plant leaves with knife & 18 & Medium \\
Nursery potting and bagging & Filling planting bags with soil; Planting seedlings in soil; Potting succulent plants into cups & 11 & Easy \\
Clearing vegetation with tools & Cutting weeds using a sickle; Weeding soil with hand tools; Clearing dry vegetation from ground & 8 & Easy \\
Hose watering and spraying & Watering plants with a hose; Hosing water or spraying pesticide; Handling and connecting water hoses & 3 & Easy \\
Manual hand weeding & Clearing weeds from soil; Weeding soil by hand; Weeding potted plants & 6 & Easy \\
Field harvesting leafy crops & Harvesting leafy crops in field; Harvesting crops from the ground; Harvesting leafy greens with shears & 4 & Easy \\
Manual soil and ground work & Digging soil with hand tool; Shoveling dirt into bags; Tilling and working soil & 8 & Easy \\
Field and orchard tending & Tending to potted plants; Carrying soil in an orchard; Tending and inspecting potted plants & 8 & Easy \\
Sickle cutting and harvesting & Cutting grass with sickle; Harvesting crops with a sickle; Cutting plant stalks with a sickle & 6 & Easy \\
Hand tool planting and harvesting & Harvesting crops with hand tools; Planting seedlings with hand tools; Harvesting root vegetables by hand & 3 & Medium \\
Cocoa and coconut processing & Bagging cocoa pods on trees; Extracting cocoa beans from pods; Processing coconut husks and shells & 4 & Medium \\
Tying and bagging plants & Wrapping and securing plant grafts; Bagging fruit on trees; Bundling fresh vegetable stems & 4 & Easy \\
\addlinespace[0.5em]
\multicolumn{4}{@{}l}{\textbf{Food Cutting and Processing} \textcolor{black!50}{(16 task types, 188 tasks)}} \\*
\cmidrule{1-4}\nopagebreak
Food assembly and plating & Prepping food ingredients on counter; Rolling and wrapping food wraps; Preparing and serving food ingredients & 28 & Medium \\
Vegetable peeling and chopping & Chopping vegetables on cutting board; Chopping ingredients on cutting board; Peeling and trimming garlic cloves & 18 & Easy \\
Banana peeling and handling & Peeling green bananas; Peeling green bananas into crates; Peeling and slicing green bananas & 16 & Easy \\
Dough portioning and shaping & Portioning and shaping dough; Shaping and arranging dough pieces; Stacking circular dough pieces & 21 & Medium \\
Plantain peeling and slicing & Peeling green plantains; Peeling green plantains into crates; Peeling plantains with knife & 10 & Easy \\
Meat butchering and slicing & Slicing and trimming raw meat; Cutting raw meat on board; Cutting meat on chopping block & 17 & Medium \\
Fish cleaning and filleting & Cleaning and gutting fish; Scaling and gutting fish; Filleting and descaling fresh fish & 11 & Medium \\
Pastry decorating and baking & Arranging baked goods on trays; Assembling and decorating cakes; Arranging dough portions on trays & 17 & Medium \\
Food weighing and bagging & Portioning and weighing dough; Weighing and packaging items; Portioning and weighing food items & 11 & Medium \\
Poultry butchering and prepping & Chopping raw poultry on block; Butchering and slicing chicken meat; Butchering and trimming raw chicken & 8 & Medium \\
Seafood and egg sorting & Peeling and deveining shrimp; Sorting eggs into trays; Arranging seafood on display trays & 11 & Easy \\
Rice cleaning and portioning & Portioning cooked rice; Washing and rinsing rice; Bagging cooked rice portions & 5 & Easy \\
Egg washing and inspection & Washing and crating duck eggs; Cleaning and inspecting eggs; Preparing and cooking egg dishes & 4 & Easy \\
Corn shelling and coconut grating & Shelling corn cobs manually; Husking and bagging corn ears; Grating fresh coconut & 6 & Easy \\
Bean bundling and shelling & Bundling long beans with bands; Shelling peas into containers & 2 & Easy \\
Machine blending and grinding & Blending ingredients in a blender; Brewing coffee with espresso equipment; Grinding ingredients with electric grinder & 3 & Easy \\
\addlinespace[0.5em]
\multicolumn{4}{@{}l}{\textbf{Manual Crafting and Assembly} \textcolor{black!50}{(16 task types, 177 tasks)}} \\*
\cmidrule{1-4}\nopagebreak
Textile weaving and crocheting & Crocheting yarn with a hook; Braiding natural fibers into rope; Weaving fabric on a loom & 33 & Hard \\
Small craft and bead assembly & Stringing beads for jewelry; Assembling decorative craft items; Preparing small craft components & 27 & Medium \\
Flower and garland arranging & Assembling floral arrangements; Weaving fresh floral garlands; Assembling flower garlands & 16 & Medium \\
Palm leaf and bamboo stripping & Stripping palm leaf midribs; Stripping palm leaves from midribs; Splitting and shaving bamboo strips & 15 & Medium \\
Surface painting and diamond art & Applying diamond painting beads; Assembling diamond painting craft; Painting structures and objects & 8 & Medium \\
Scissor cutting and trimming & Cutting paper into strips; Cutting paper into containers; Cutting craft materials with scissors & 12 & Easy \\
Handicraft assembly and small packaging & Packaging small items into plastic; Assembling small handmade crafts; Crafting items with pipe cleaners & 16 & Easy \\
Craft table organization and assembly & Assembling handmade craft decorations; Unpacking and organizing desk items; Preparing craft materials on table & 12 & Easy \\
Workbench assembly and tool setup & Assembling small components manually; Setting up workstation and assembling; Threading hair for clients & 11 & Medium \\
Footwear crafting and repair & Repairing shoes by hand stitching; Stitching shoe soles by hand; Operating shoe component pressing machine & 5 & Hard \\
Seated crafting and domestic chores & Resting and taking breaks; Manipulating small craft objects; Handling small materials while seated & 7 & Easy \\
Sculpture and costume fabrication & Sanding sculpted figures; Assembling costume headpieces; Carving detailed clay sculpture & 7 & Hard \\
Garment and gear adjustments & Adjusting garments and clothing; Adjusting workstation items and tools; Assembling and inspecting straps & 4 & Easy \\
Bamboo lantern assembly & Assembling bamboo star lanterns & 1 & Hard \\
Badminton racket restringing & Restringing a badminton racket & 1 & Hard \\
Carton assembly and packing & Organizing and preparing craft containers; Assembling and packing cartons & 2 & Easy \\
\addlinespace[0.5em]
\multicolumn{4}{@{}l}{\textbf{Packaging and Material Bagging} \textcolor{black!50}{(15 task types, 190 tasks)}} \\*
\cmidrule{1-4}\nopagebreak
Small parts bagging and sealing & Bagging small parts; Packing items into plastic packaging; Filling and preparing plastic bags & 27 & Easy \\
Cardboard box folding and packing & Folding and packing small boxes; Assembling and packaging matchboxes; Packing items into cardboard boxes & 19 & Easy \\
Food bagging and portioning & Packing snacks into plastic bags; Bagging food items into plastic bags; Bagging food and sauces & 15 & Easy \\
Candy shaping and packaging & Packaging candy canes into bags; Pulling and stretching candy strips; Pulling and cutting candy strips & 13 & Medium \\
Paper and film wrapping & Wrapping items in paper packaging; Folding paper sheets and boxes; Wrapping objects in plastic film & 17 & Easy \\
Retail counter bagging and packing & Bagging items at retail counter; Preparing small merchandise packets; Organizing merchandise on a counter & 19 & Easy \\
Portioning and small packet filling & Preparing plastic packaging material; Packing materials into small packets; Preparing small packages at table & 13 & Easy \\
Packaging plates in plastic sleeves & Packaging items in plastic sleeves; Packaging paper plates in plastic; Packaging round items into wrappers & 9 & Easy \\
Operating plate press machines & Operating plate forming press; Operating a container press machine; Operating a press machine & 10 & Medium \\
Floor-based sorting and packing & Packing and wrapping packages on floor; Sorting materials on floor; Arranging packaged items on floor & 10 & Easy \\
Tray arranging and case packing & Arranging food portions on trays; Arranging clay lamps in trays; Assembling craft pieces on trays & 11 & Easy \\
Sorting and stacking plates & Inspecting and stacking paper plates; Sorting and stacking leaf plates; Sorting and stacking paper plates & 4 & Easy \\
Workstation packing and material handling & Packing and labeling items; Packaging small items at workstation; Packing merchandise at workstation & 11 & Medium \\
Stocking and organizing shelves & Organizing packaged goods on shelves; Sorting and organizing small sachets; Organizing items into partitioned containers & 8 & Easy \\
Medication packing and dispensing & Dispensing medication blister packs; Sorting and packing medication boxes; Retrieving tools and packaged items & 4 & Easy \\
\addlinespace[0.5em]
\multicolumn{4}{@{}l}{\textbf{Textile and Garment Finishing} \textcolor{black!50}{(14 task types, 120 tasks)}} \\*
\cmidrule{1-4}\nopagebreak
Cutting fabric with scissors & Cutting fabric with scissors; Trimming fabric pieces with scissors; Cutting and trimming fabric pieces & 21 & Medium \\
Trimming loose threads & Trimming loose threads from garments; Trimming loose garment threads; Trimming loose threads on garments & 13 & Easy \\
Preparing fabric and upholstery & Inspecting and folding fabric pieces; Arranging fabric layers on table; Sorting and arranging fabric scraps & 18 & Medium \\
Hand washing laundry & Washing clothes by hand; Washing laundry by hand; Hand washing clothes on washboard & 10 & Easy \\
Inspecting and marking fabrics & Inspecting and marking fabric pieces; Inspecting and trimming garments; Inspecting and trimming garment threads & 13 & Medium \\
Folding garments and clothes & Folding clothes and garments; Folding white garments; Folding garments and fabrics & 10 & Easy \\
Pressing fabric and mending & Pressing fabric with an iron; Ironing clothes and fabric; Ironing and folding garments & 5 & Medium \\
Ironing garments on board & Ironing garments on ironing board; Ironing shirts on a board; Ironing garments on a board & 7 & Easy \\
Organizing and packing laundry & Folding and organizing laundry; Folding laundry on a bed; Folding and organizing clothing & 5 & Easy \\
Multi-step garment finishing & Ironing and pressing garments; Folding garments on a table; Folding and trimming white garments & 8 & Easy \\
Ironing shirts and clothes & Ironing garments on board; Ironing shirts and garments; Ironing and folding clothing items & 3 & Medium \\
Hanging laundry to dry & Hanging laundry to dry; Hanging clothes on hangers; Hanging laundry on clothesline & 3 & Easy \\
Inspecting and folding garments & Inspecting and folding fabric garments; Inspecting and folding garments & 2 & Easy \\
Assembling footwear and socks & Assembling footwear components; Folding socks and laundry & 2 & Easy \\
\addlinespace[0.5em]
\multicolumn{4}{@{}l}{\textbf{Cleaning and Sanitation} \textcolor{black!50}{(19 task types, 152 tasks)}} \\*
\cmidrule{1-4}\nopagebreak
Sink and basin dishwashing & Washing dishes in the sink; Washing dishes by hand; Washing dishes in basins & 22 & Easy \\
Wiping bottled goods and surfaces & Wiping cans with a cloth; Wiping bottles with cloth; Wiping surfaces with a cloth & 14 & Easy \\
Indoor floor mopping and sweeping & Sweeping floor with broom; Cleaning floors with vacuum; Mopping the floor & 13 & Easy \\
Sweeping outdoor grounds and streets & Sweeping ground with a broom; Sweeping roadside debris with broom; Sweeping debris into dustpan & 7 & Easy \\
Cooking and kitchen sink cleanup & Washing dishes in a sink; Cleaning utensils and washing hands; Washing dishes and utensils & 9 & Easy \\
Cleaning retail and kitchen counters & Wiping and organizing canned goods; Wiping down kitchen countertops; Cleaning and organizing service counters & 13 & Easy \\
Washing water jugs and containers & Washing large water containers; Washing plastic water jugs; Washing items in a basin & 8 & Easy \\
Scrubbing cookware and food prep & Cleaning plastic water containers; Washing dishes under running water; Scrubbing and cleaning pots & 10 & Easy \\
Cleaning machined parts and scrap & Cleaning printed circuit boards; Cleaning metal structures and objects; Cleaning and inspecting machined parts & 10 & Medium \\
Refilling and capping water jugs & Refilling water jugs; Dispensing water from dispenser; Refilling and capping water jugs & 10 & Easy \\
Washing produce dishes and kitchenware & Washing dishes and containers; Washing ingredients and dishes; Washing and rinsing food items & 8 & Easy \\
Bussing and cleaning dining tables & Wiping and cleaning table; Cleaning and organizing tables; Cleaning food preparation station & 6 & Easy \\
Sponge washing dishes and cookware & Washing dishes with a sponge; Washing cookware and utensils; Washing surfaces with a sponge & 3 & Easy \\
Maintaining doors and window frames & Cleaning window frames and grilles; Wiping door and frame surfaces; Installing and finishing door frames & 6 & Easy \\
Cleaning office desks and workstations & Cleaning office desk surfaces; Cleaning workstation surfaces & 2 & Easy \\
Washing and organizing dish racks & Washing and organizing dishes; Washing dishes and placing racks; Organizing dishes and kitchenware & 4 & Easy \\
Cleaning and repairing electric fans & Cleaning fan parts with cloth; Cleaning equipment using compressed air; Repairing and adjusting electric fans & 3 & Easy \\
Pumping and carrying water buckets & Pumping and transferring water; Retrieving supplies from buckets; Carrying water buckets across yard & 3 & Easy \\
Scraping and sweeping outdoor pavement & Sweeping and scraping outdoor surfaces & 1 & Easy \\
\addlinespace[0.5em]
\multicolumn{4}{@{}l}{\textbf{Metalworking and Heavy Machining} \textcolor{black!50}{(11 task types, 106 tasks)}} \\*
\cmidrule{1-4}\nopagebreak
Welding and metal assembly & Welding metal structures; Assembling hardware components; Assembling metal frame structures & 22 & Hard \\
Conventional milling and machining & Operating industrial milling machines; Assembling and adjusting metal frames; Operating and adjusting lathe machine & 19 & Hard \\
Angle grinder metal finishing & Grinding metal with angle grinder; Grinding metal parts and joints; Grinding and shaping metal parts & 7 & Hard \\
Rod and cane frame bending & Bending rebar and metal rods; Constructing bamboo frame crafts; Bending cane poles with jig & 12 & Hard \\
Workshop machining and woodworking & Assembling mechanical and hardware components; Measuring and marking boards; Machining and inspecting metal components & 13 & Medium \\
Hot metal anvil forging & Forging hot metal on an anvil; Heating metal in a forge; Forging heated metal on anvil & 7 & Hard \\
CNC operation and machine monitoring & Operating industrial processing machinery; Observing the work area; Observing surroundings while resting & 11 & Easy \\
Workpiece measuring and marking & Measuring and marking sheet metal; Measuring and marking metal pieces; Fabricating and inspecting metal frames & 6 & Hard \\
Power cutting and chiseling & Cutting metal with an angle grinder; Grinding surfaces with power tool; Cutting stone slabs with power cutter & 4 & Hard \\
Precision measurement and alignment & Measuring workpieces with precision instruments; Measuring materials with tape measure; Aligning workpieces with dial indicator & 3 & Hard \\
Manual blade sharpening & Sharpening blades and knives; Sharpening blade with whetstone & 2 & Hard \\
\addlinespace[0.5em]
\multicolumn{4}{@{}l}{\textbf{Woodworking and Carpentry} \textcolor{black!50}{(10 task types, 91 tasks)}} \\*
\cmidrule{1-4}\nopagebreak
Wood sawing and cutting & Trimming and sharpening thin sticks; Sawing wood with a handsaw; Cutting wooden planks with a saw & 19 & Hard \\
Lumber milling and preparation & Stacking wooden planks; Sorting and preparing wooden planks; Measuring and marking wooden boards & 19 & Medium \\
Wood layout marking and sketching & Measuring and marking wooden boards; Measuring and marking boards; Drawing sketch on wooden board & 12 & Medium \\
Hand wood carving and whittling & Carving wood with hand tools; Assembling wooden framework; Whittling wooden sticks with knives & 13 & Hard \\
Frame nailing and fastening & Assembling wooden frames with nailgun; Assembling wooden frame with nailer; Assembling wooden frames with hammer & 12 & Medium \\
Workshop furniture and frame fabrication & Assembling wooden frame structures; Carrying wooden boards across workshop; Assembling small hardware components & 6 & Hard \\
Wood sanding and surface cleaning & Sanding wooden boards; Sanding and cleaning wooden boards; Wiping whiteboard and desk surfaces & 4 & Medium \\
Bed frame slat assembly & Assembling wooden bed frame slats; Arranging and processing wooden slats & 2 & Medium \\
Miniature wood model assembly & Assembling miniature wooden model pieces; Assembling wooden frame with slats; Assembling wooden frames with nailgun & 3 & Hard \\
Wooden crate assembly & Assembling wooden crate with nailer & 1 & Medium \\
\addlinespace[0.5em]
\multicolumn{4}{@{}l}{\textbf{Retail Sales and Inventory} \textcolor{black!50}{(15 task types, 119 tasks)}} \\*
\cmidrule{1-4}\nopagebreak
Produce weighing and bagging & Packing vegetables into plastic bags; Weighing and packaging produce; Weighing and bagging customer produce & 11 & Medium \\
Shipment unpacking and product inspection & Inspecting products on retail shelves; Unpacking and inspecting retail shipments; Inspecting grocery items on shelves & 12 & Easy \\
Market produce trimming and display & Arranging fresh produce on display; Trimming and bagging market vegetables; Sorting produce at market stall & 13 & Easy \\
Cash counting and ledger calculation & Counting and organizing paper currency; Calculating expenses and logging records; Calculating numbers and counting cash & 10 & Medium \\
Market stall produce vending & Weighing produce and serving customers; Weighing and bagging stall produce; Managing cash transactions at stall & 13 & Easy \\
Counter cashier and sales management & Handling cash transactions at counter; Managing cash in a drawer; Serving customers at counter & 9 & Medium \\
Customer packaging and checkout & Weighing and selling produce; Packaging items and serving customers; Serving customers and handling cash & 10 & Medium \\
Pharmacy medication dispensing & Dispensing blister pack medications; Dispensing medication to customers; Dispensing medication at pharmacy counter & 8 & Hard \\
Fruit display and restocking & Arranging fruit in produce display; Restocking store display shelves; Packing produce into plastic trays & 6 & Easy \\
Snack rack and hook stocking & Organizing snacks on display racks; Stocking snack packages on displays; Hanging snack bags on display racks & 7 & Easy \\
General store attending and browsing & Browsing supermarket grocery shelves; Managing store counter and paperwork; Packaging goods into plastic bags & 7 & Easy \\
Product and bottle inspection & Inspecting retail products on shelves; Inspecting bottles on conveyor line; Conducting laboratory titration & 5 & Hard \\
Retail shelf inventory inspection & Inspecting retail products on shelves & 1 & Easy \\
Inventory logbook and ledger recording & Recording inventory in a notebook; Reviewing and filing technical documents; Recording entries in ledger book & 4 & Medium \\
Counter downtime and note taking & Leaving the service counter; Prepping produce and stall items; Writing notes and managing paperwork & 3 & Easy \\
\addlinespace[0.5em]
\multicolumn{4}{@{}l}{\textbf{Digital Interaction and Workplace Tasks} \textcolor{black!50}{(9 task types, 84 tasks)}} \\*
\cmidrule{1-4}\nopagebreak
Workplace smartphone operation & Operating a smartphone; Using a smartphone; Operating smartphone while seated & 15 & Easy \\
Desktop computer workstation tasks & Operating a computer workstation; Editing photos on a computer; Operating a desktop computer & 11 & Easy \\
Paperwork and record keeping & Reviewing and calculating paperwork; Writing notes on paper pad; Writing notes at counter & 15 & Medium \\
Food stall smartphone multitasking & Operating smartphone at counter; Prepping food ingredients at stall; Food preparation and phone multitasking & 10 & Easy \\
Food preparation and craft assembly & Stepping away from prep table; Crafting and assembling decorative pieces; Operating equipment and stepping away & 9 & Medium \\
Casual and break smartphone browsing & Operating a smartphone at table; Walking while using a smartphone; Using smartphone during work breaks & 7 & Easy \\
Mobile device repair and inspection & Using a smartphone; Trimming patches and using phone; Repairing mobile phone devices & 6 & Easy \\
Coworker interaction and collaboration & Assembling structures with coworkers; Conversing with workplace colleagues; Interacting with coworkers during tasks & 7 & Easy \\
Outdoor tool and material transport & Walking outdoors along paths; Carrying items outdoors; Handling hand tools outdoors & 4 & Easy \\
\addlinespace[0.5em]
\multicolumn{4}{@{}l}{\textbf{Cooking and Culinary Tasks} \textcolor{black!50}{(9 task types, 105 tasks)}} \\*
\cmidrule{1-4}\nopagebreak
Skewering grilling and deep frying & Skewering marinated food pieces; Stir-frying ingredients in a wok; Grilling food skewers over charcoal & 19 & Medium \\
Stovetop pot and pan cooking & Cooking food on the stovetop; Cooking food in a pan; Cooking food in small pan & 19 & Medium \\
Market stall food vending & Preparing food at a market stall; Handling money and stall transactions; Organizing goods at a stall & 16 & Medium \\
Food plating and portioning & Eating a meal; Portioning food into containers; Plating and serving food & 14 & Easy \\
Food station ingredient preparation & Prepping ingredients in the kitchen; Prepping food ingredients in kitchen; Preparing food in kitchen & 11 & Medium \\
Beverage preparation and service & Preparing blended and iced drinks; Preparing iced beverages at counter; Crafting with plastic bottles & 9 & Medium \\
Cold storage and food packaging & Filling ice candy wrappers; Stocking and retrieving refrigerated items; Retrieving ingredients from refrigerator & 7 & Easy \\
Kitchen and dining navigation & Walking through a kitchen area; Cooking food on a stovetop; Working in a kitchen & 7 & Easy \\
Noodle preparation and sauce bottling & Cooking and bottling sauces; Cooking and preparing noodles; Cooking and portioning noodles & 3 & Medium \\
\addlinespace[0.5em]
\multicolumn{4}{@{}l}{\textbf{Plastics and Pipe Fabrication} \textcolor{black!50}{(6 task types, 56 tasks)}} \\*
\cmidrule{1-4}\nopagebreak
Handsaw plastic pipe cutting & Cutting plastic pipes with saw; Cutting plastic pipes with handsaw; Sorting and cutting scrap pipes & 14 & Easy \\
Pipe wrapping and plumbing assembly & Unwrapping plastic film from pipes; Wrapping pipes with plastic film; Wrapping pipes with protective film & 11 & Medium \\
Pipe cleaner crafting and sorting & Making pipe cleaner crafts; Sorting scrap plastic pipes; Sorting and cutting plastic pipes & 4 & Easy \\
Bulk plastic shredding and bagging & Processing scrap plastic pieces; Collecting loose materials into sacks; Transferring items between containers & 13 & Easy \\
Pipe cleaning trimming and carving & Cleaning and trimming plastic pipes; Carving and detailing wooden pipes; Cutting scrap pipes & 6 & Medium \\
Bottle handling stocking and sorting & Wiping and arranging shelf bottles; Crafting objects from plastic bottles; Handling a water bottle & 8 & Easy \\
\addlinespace[0.5em]
\multicolumn{4}{@{}l}{\textbf{Furniture Assembly and Upholstery} \textcolor{black!50}{(9 task types, 69 tasks)}} \\*
\cmidrule{1-4}\nopagebreak
Cushion and chair frame upholstery & Upholstering furniture cushions; Upholstering furniture frame; Upholstering furniture with fabric & 18 & Hard \\
Pneumatic stapler furniture upholstery & Upholstering furniture with staple gun; Upholstering frame using pneumatic stapler; Stapling fabric onto furniture components & 10 & Hard \\
Structural frame and headboard assembly & Upholstering wooden furniture frame; Refurbishing and gripping cricket bats; Assembling decorative craft pieces & 12 & Hard \\
Adhesive and filler application & Applying adhesive and stencil resin; Applying adhesive or finish to wood; Applying putty with spatulas & 7 & Medium \\
Fabric lining and fastener attachment & Upholstering fabric to furniture frame; Stapling fabric lining to furniture; Handling and arranging black fabric & 6 & Medium \\
Foam padding cutting and shaping & Attaching foam padding to frames; Cutting and shaping foam components; Trimming foam padding for upholstery & 6 & Medium \\
Board preparation and rope crafting & Attaching rope handles to boards; Planing and marking wooden boards; Processing materials on wooden boards & 5 & Medium \\
Coffin interior lining upholstery & Upholstering coffin interior with fabric; Upholstering casket interior lining; Constructing and upholstering wooden coffins & 3 & Hard \\
Casket hardware and trim installation & Attaching decorative trim to coffins; Installing hardware on wooden caskets & 2 & Hard \\
\addlinespace[0.5em]
\multicolumn{4}{@{}l}{\textbf{Vehicle Maintenance and Repair} \textcolor{black!50}{(7 task types, 76 tasks)}} \\*
\cmidrule{1-4}\nopagebreak
Motorcycle engine and component maintenance & Servicing motorcycle engine components; Cleaning motorcycle engine parts; Inspecting and maintaining motorcycle components & 33 & Hard \\
Bicycle assembly and tire repair & Assembling bicycle frames and parts; Lacing bicycle wheel spokes; Adjusting and repairing bicycle components & 8 & Hard \\
Auto body detailing and prepping & Sanding vehicle body panels; Buffing and polishing a car; Masking vehicle parts with tape & 10 & Hard \\
Precision engine and electromechanical disassembly & Servicing motorcycle brake calipers; Disassembling electromechanical appliances; Servicing engine cylinder head components & 9 & Hard \\
Scooter mechanical repair and servicing & Maintaining and servicing scooters; Dismantling and repairing scooter parts; Servicing scooter mechanical parts & 6 & Hard \\
Vehicle frame and chassis assembly & Assembling vehicle frame components; Inspecting and repairing vehicle components; Fastening machine components with tools & 6 & Hard \\
Vehicle washing and exterior cleaning & Washing vehicle wheels and exterior; Wiping down vehicle bodywork; Washing vehicle wheels and mats & 4 & Easy \\
\addlinespace[0.5em]
\multicolumn{4}{@{}l}{\textbf{Personal Care and Animal Services} \textcolor{black!50}{(4 task types, 37 tasks)}} \\*
\cmidrule{1-4}\nopagebreak
Barbering and hair cutting & Cutting client hair with clippers; Cutting hair with scissors and clippers; Shaving face with straight razor & 10 & Hard \\
Pet dog grooming & Grooming dog coat with clippers; Grooming dog facial fur; Brushing and grooming a dog & 14 & Hard \\
Manicure and pedicure services & Filing client fingernails for manicure; Performing a pedicure on client; Trimming and filing nails & 6 & Medium \\
Animal care and pen cleaning & Washing pigs with a hose; Cleaning the cat litter box; Grooming cat ears and fur & 7 & Easy \\
\addlinespace[0.5em]
\multicolumn{4}{@{}l}{\textbf{Electronics and Circuit Assembly} \textcolor{black!50}{(5 task types, 51 tasks)}} \\*
\cmidrule{1-4}\nopagebreak
Circuit board soldering and assembly & Soldering components onto circuit board; Assembling and inspecting circuit boards; Soldering components on circuit boards & 18 & Hard \\
Mechanical parts inspection and finishing & Measuring and inspecting metal parts; Assembling and inspecting small components; Inspecting and cleaning mechanical parts & 11 & Medium \\
Wire and cable assembly & Assembling wiring and cable components; Wiring and soldering electrical components; Wiring and taping electrical connections & 11 & Medium \\
Panel edge trimming and assembly & Trimming and deburring panel edges; Trimming board edge banding; Trimming circuit board leads & 6 & Medium \\
Unpacking and sorting goods & Unpacking and sorting boxed goods; Unpacking and organizing components; Unpacking items from plastic packaging & 5 & Easy \\
\addlinespace[0.5em]
\multicolumn{4}{@{}l}{\textbf{Building Construction and Masonry} \textcolor{black!50}{(7 task types, 45 tasks)}} \\*
\cmidrule{1-4}\nopagebreak
Masonry plastering and tiling & Applying mortar to masonry wall; Applying and smoothing wall plaster; Sanding wall and ceiling surfaces & 17 & Hard \\
Mortar mixing and concrete finishing & Applying mortar between slabs; Mixing and applying mortar; Mixing cement mortar with hoe & 9 & Hard \\
Rebar assembly and scaffolding & Tying rebar intersections with wire; Tying steel rebar reinforcement cages; Fabricating and assembling rebar cages & 4 & Hard \\
Demolition excavation and metalwork & Welding and installing staircase railings; Digging trenches in the soil; Breaking concrete with demolition hammer & 6 & Medium \\
Interior wall painting & Painting interior wall with brush; Painting walls with roller; Painting ceiling with roller & 3 & Medium \\
Masonry and construction site labor & Transporting materials across the workspace; Masonry work on a rooftop; Walking through construction site & 5 & Easy \\
Wheelbarrow material transport & Transporting materials with a wheelbarrow & 1 & Easy \\
\end{longtable}

\clearpage
\section{Hand Pose Gallery}
\label{sec:supp_hand_gallery}
\begin{center}
\captionsetup{type=figure}
\includegraphics[width=\textwidth]{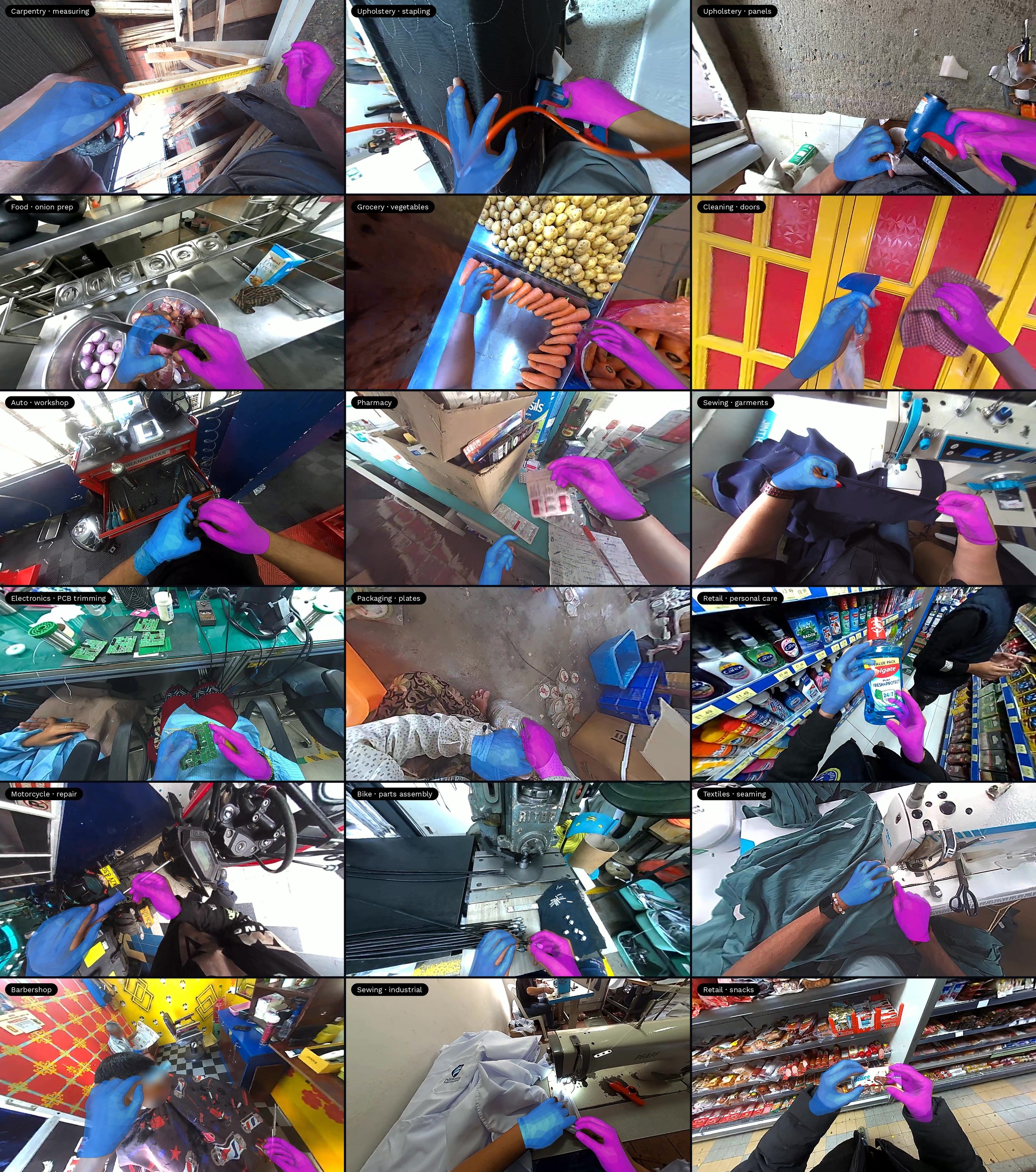}
\caption{\textbf{Hand pose estimates across workplaces (1/2).} Thirty-six further sessions extending Figure~\ref{fig:ego_hands}, one frame each, chosen to span workplaces, tools and hand--object contact. Estimated MANO hand meshes are rendered over egocentric frames (blue: left hand, magenta: right hand).}
\label{fig:supp_hands_1}
\end{center}
\clearpage
\begin{center}
\captionsetup{type=figure}
\includegraphics[width=\textwidth]{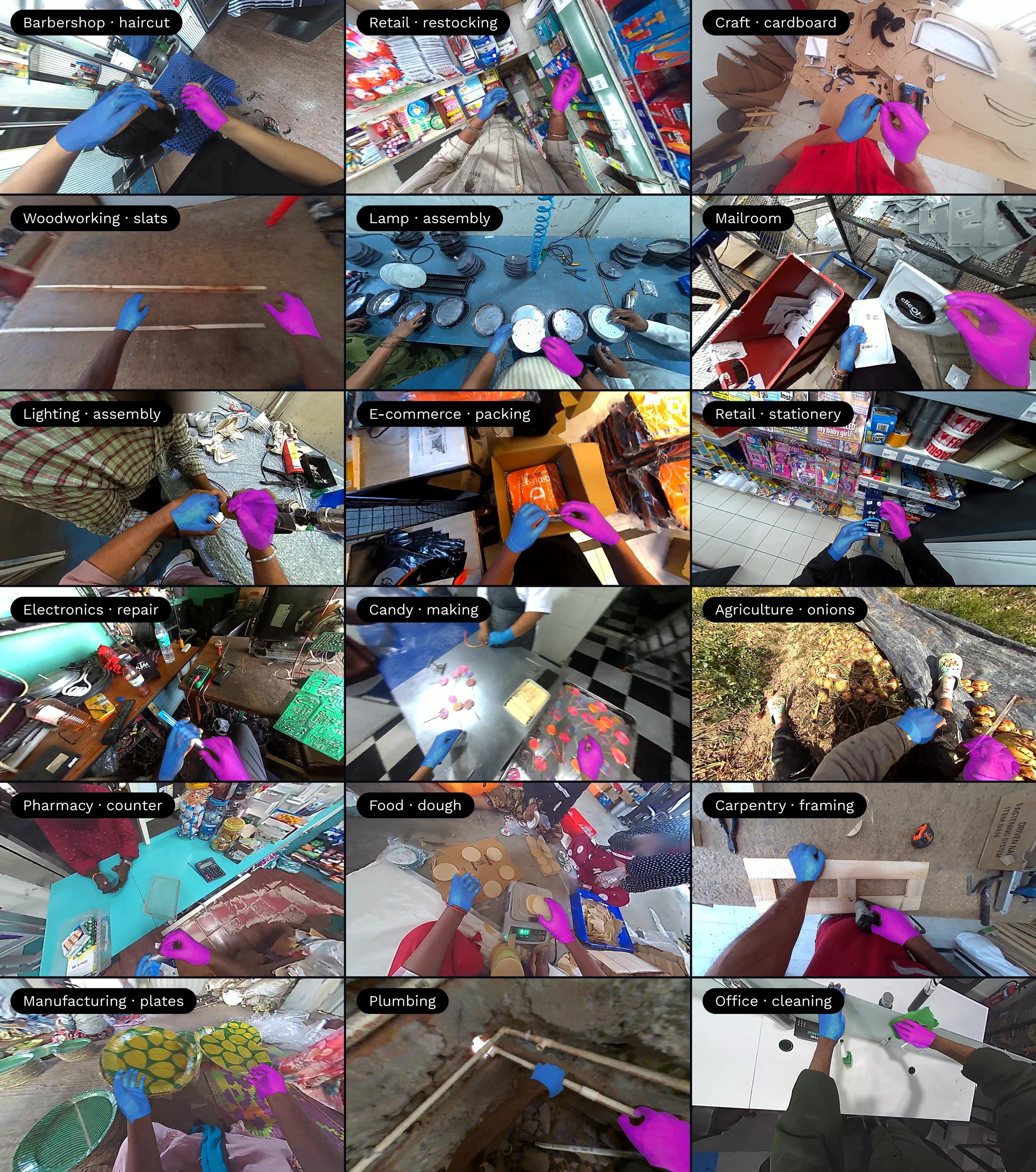}
\caption{\textbf{Hand pose estimates across workplaces (2/2).} Continued from Figure~\ref{fig:supp_hands_1}.}
\label{fig:supp_hands_2}
\end{center}

\clearpage
\section{Face Blurring}

\begin{figure}[H]
\centering
\begin{minipage}[t]{0.495\textwidth}
  \centering
  \includegraphics[width=\linewidth]{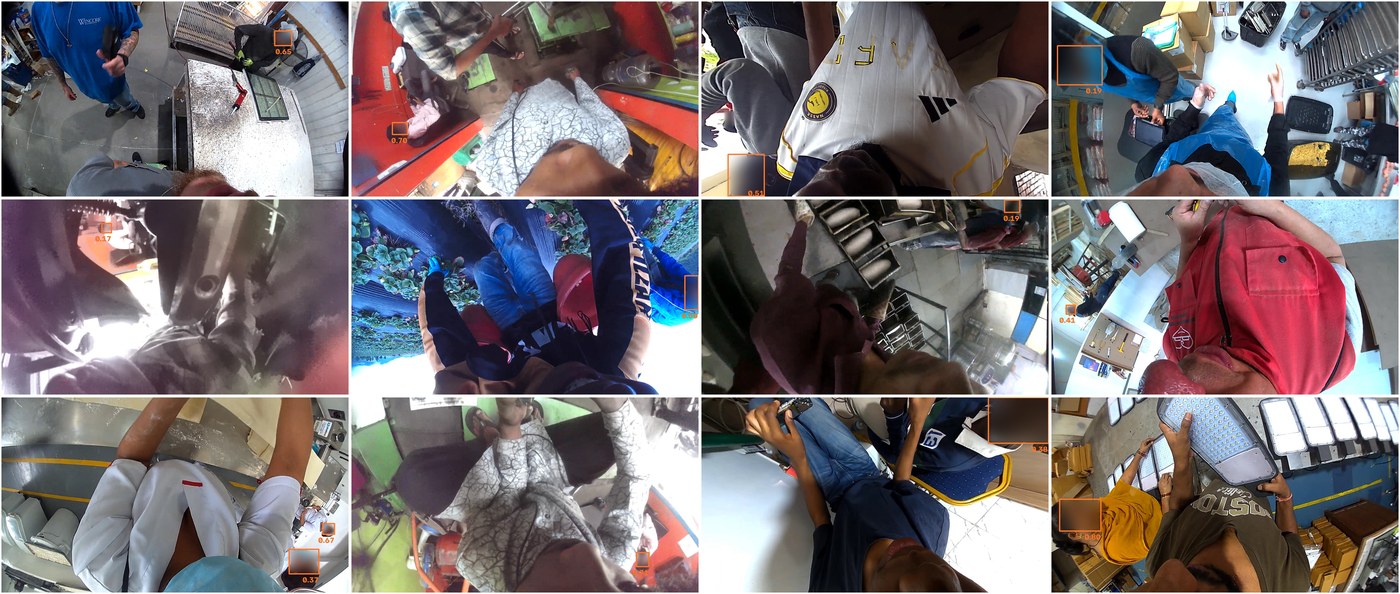}\\[2pt]
  {\footnotesize (a) Eye cameras}
\end{minipage}\hfill
\begin{minipage}[t]{0.495\textwidth}
  \centering
  \includegraphics[width=\linewidth]{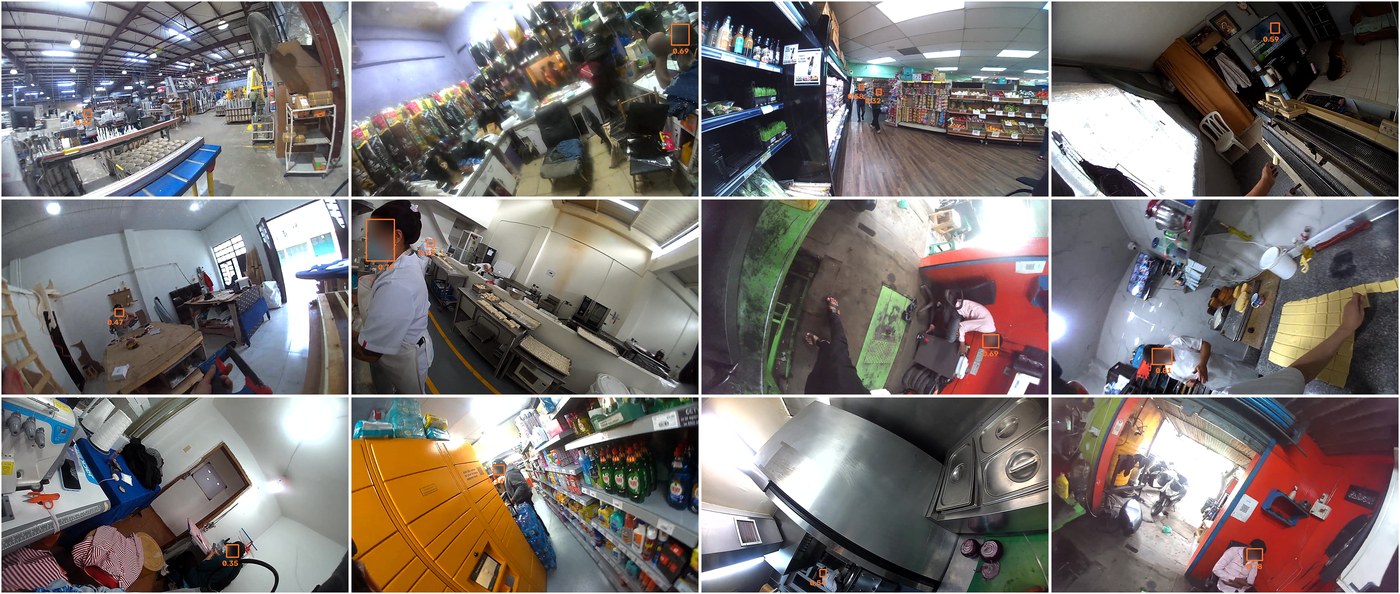}\\[2pt]
  {\footnotesize (b) Lateral cameras}
\end{minipage}
\caption{\textbf{Additional face anonymization examples} from the eye and lateral cameras, where faces are strongly distorted or oriented.}
\label{fig:supp_face_examples}
\end{figure}

\end{document}